\documentclass{article}

\usepackage{microtype}
\usepackage{graphicx}
\usepackage{subcaption}
\usepackage{booktabs} 

\usepackage{hyperref}

\usepackage[preprint]{icml2026}

\usepackage{amsmath}
\usepackage{amssymb}
\usepackage{mathtools}
\usepackage{amsthm}
\usepackage{enumitem}

\usepackage[table]{xcolor}

\usepackage[capitalize,noabbrev]{cleveref}

\theoremstyle{plain}

\theoremstyle{definition}

\theoremstyle{remark}

\usepackage[textsize=tiny]{todonotes}

\icmltitlerunning{SENDAI for Data Assimilation}

\begin{document}

\twocolumn[
  \icmltitle{SENDAI: A Hierarchical Sparse-measurement, EfficieNt Data AssImilation Framework}




  \icmlsetsymbol{equal}{*}

  \begin{icmlauthorlist}
    \icmlauthor{Xingyue Zhang}{equal,env}
    \icmlauthor{Yuxuan Bao}{equal,amath}
    \icmlauthor{Mars Liyao Gao}{cse}
    \icmlauthor{J. Nathan Kutz}{amath,ece}

  \end{icmlauthorlist}

  \icmlaffiliation{ece}{Department of Electrical and Computer Engineering, University of Washington, Seattle, USA}
  \icmlaffiliation{cse}{Paul G. Allen School of Computer Science \& Engineering, University of Washington, Seattle, USA}
  \icmlaffiliation{amath}{Department of Applied Mathematics, University of Washington, Seattle, USA} 
  \icmlaffiliation{env}{School of Environmental and Forest Sciences, University of Washington, Seattle, USA}

  \icmlcorrespondingauthor{J. Nathan Kutz}{kutz@uw.edu}

  \icmlkeywords{Machine Learning, ICML, Rotating detonation engines, data assimilation, SHRED, latent dynamics, surrogate modeling, injector physics}

  \vskip 0.3in
]



\printAffiliationsAndNotice{\icmlEqualContribution}

\begin{abstract}

Bridging the gap between data-rich training regimes and observation-sparse deployment conditions remains a central challenge in spatiotemporal field reconstruction, particularly when target domains exhibit distributional shifts, heterogeneous structure, and multi-scale dynamics absent from available training data. We present SENDAI, a hierarchical $\textbf{S}$parse-measurement, $\textbf{E}$fficie$\textbf{N}$t $\textbf{D}$ata $\textbf{A}$ss$\textbf{I}$milation Framework that reconstructs full spatial states from hyper sparse sensor observations by combining simulation-derived priors with learned discrepancy corrections. We demonstrate the performance on satellite remote sensing, reconstructing MODIS (Moderate Resolution Imaging Spectroradiometer) derived  vegetation index fields across six globally distributed sites. Using seasonal periods as a proxy for domain shift, the framework consistently outperforms established baselines that require substantially denser observations---SENDAI achieves a maximum SSIM improvement of $185\%$ over traditional baselines and a $36\%$ improvement over recent high-frequency-based methods. These gains are particularly pronounced for landscapes with sharp boundaries and sub-seasonal dynamics; more importantly, the framework effectively preserves diagnostically relevant structures---such as field topologies, land cover discontinuities, and spatial gradients. By yielding corrections that are more structurally and spectrally separable, the reconstructed fields are better suited for downstream inference of indirectly observed variables. The results therefore highlight a lightweight and operationally viable framework for sparse-measurement reconstruction that is applicable to physically grounded inference, resource-limited deployment, and real-time monitor and control.

\end{abstract}


\begin{figure}[t]
    \centering
    \includegraphics[width=\columnwidth]{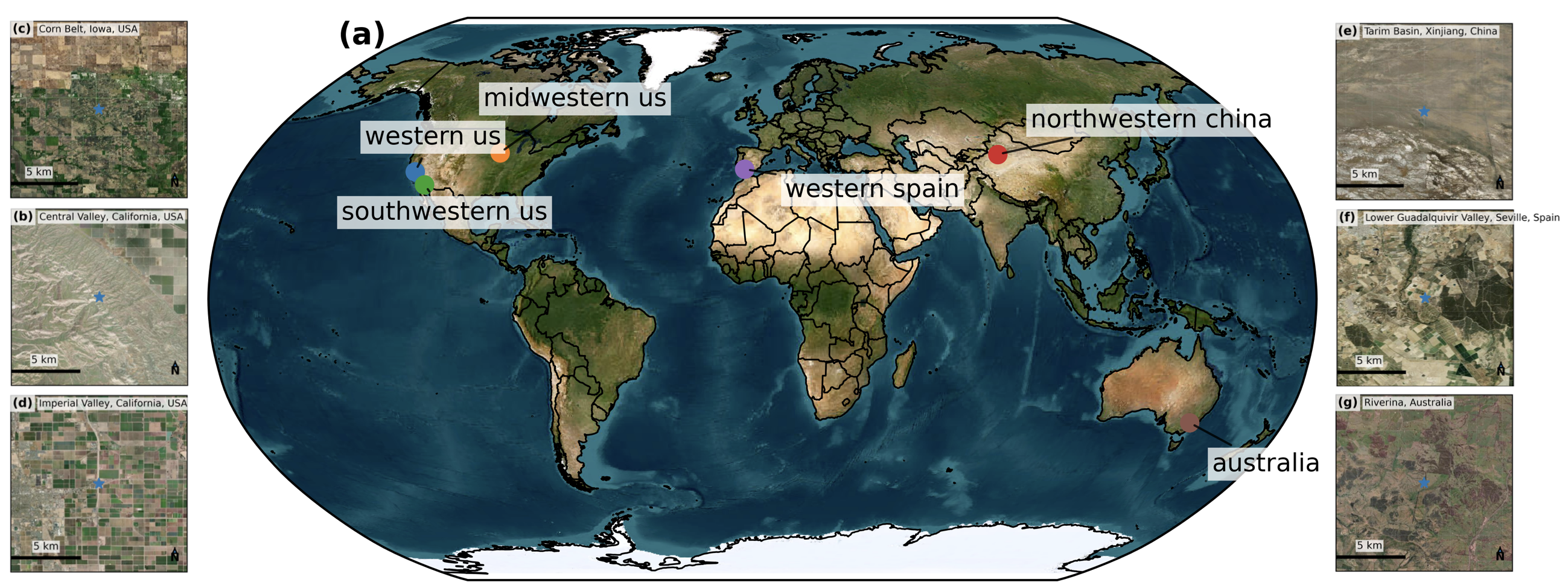}
    \caption{Study sites for NDVI reconstruction experiments. (a) Global distribution of the six study sites, spanning: (b, f) Mediterranean, (c) continental, (d, e) arid, and (g) subtropical climates.}
    \vspace{-4mm}
    \label{fig:study_sites_world}
\end{figure}

\section{Introduction}
\label{sec:intro}

Reconstructing full spatiotemporal fields from sparse observations constitutes a fundamental challenge across Earth sciences, with applications spanning vegetation monitoring, hydrological modeling, and climate analysis~\citep{weiss2020remote, mohanty2017soil,adrian2025data, jiang2025hierarchical}. Satellite remote sensing platforms such as MODIS provide unprecedented global coverage, yet cloud contamination, sensor gaps, and transmission constraints frequently yield incomplete spatial fields that compromise downstream analyses~\citep{shen2015missing, zhang2018missing}. Traditional methods typically require substantial observational coverage to achieve high-fidelity reconstruction~\citep{stock2020comparison}.

Contemporary deep learning approaches have demonstrated impressive reconstruction capabilities but typically demand GPU clusters, massively labeled training datasets, and substantial computational resources that preclude deployment in operational or resource-constrained settings~\citep{he2016identity,morel2025predicting, meraner2020cloud, cresson2018framework, zhang2018missing}, while still exhibiting considerable data dependency~\citep{sarafanov2020machine}. This computational burden is particularly problematic for near-real-time agricultural monitoring, where latency constraints and bandwidth limitations necessitate lightweight yet accurate reconstruction~\citep{gao2021mapping, denby2020orbital}. Recent works in physics-informed AI~\citep{kutz2025accelerating, karniadakis2021physics, raissi2019physics,fan2025physics,liu2024kan} and neural operators~\cite{lu2021learning, li2020fourier, roy2025physics} have shown how physical structure can be embedded into surrogates for improved generalization. However, many practical reconstruction problems are not naturally posed as a well-specified PDE with reliable priors at the spatial and temporal scales of interest, motivating methods that remain effective under weak or unknown governing structure. Moreover, the assumption of stationarity in the underlying field distribution is frequently violated in Earth observation contexts where phenological shifts, seasonal transitions, and land cover dynamics introduce substantial domain changes over time~\citep{zeng2024unsupervised, truong2021bi, cheng2025sparse}. 

Additionally, existing approaches often learn latent representations that blend multiple contributing effiects, yielding entangled corrections with limited physical interpretability~\citep{dylewsky2019dynamic, wang2022ams, chen2022automated}. This entanglement hinders inverse inference of indirectly observed fields (e.g., soil moisture, land surface temperature)---an essential capability in environmental monitoring and other Earth-science settings where target quantities are sparsely observed, intermittently available, or infeasible to measure directly at scale~\citep{koronaki2025disentangling}.

In this work, we present SENDAI (\textbf{S}parse-measurement, $\textbf{E}$fficie$\textbf{N}$t \textbf{D}ata \textbf{A}ss\textbf{I}milation), a hierarchical data assimilation framework that reconstructs full spatial states from severely sparse sensor observations by combining simulation-derived priors with learned discrepancy corrections. The architecture decomposes reconstruction into two complementary pathways: (i) a \emph{low-frequency pathway} that leverages Takens' embedding theorem~\citep{takens2006detecting} through shallow recurrent decoder networks (SHRED)~\citep{williams2024sensing} to capture dominant spatiotemporal dynamics, with latent-space adversarial alignment bridging distribution shifts between simulation and ground truth; and (ii) a \emph{high-frequency pathway} employing sequential frequency peeling with coordinate-based implicit neural representations (INRs)~\citep{sitzmann2020implicit, tancik2020fourier} to resolve fine-scale structure, sharp boundaries, and localized corrections that smooth decoders cannot capture. This hierarchical decomposition enables the framework to address heterogeneous spatiotemporal fields exhibiting domain changes, topological variations, and multi-scale structure~\citep{barwey2025mesh}.  Moreover, it overcomes the limitations of spectral bias and band-pass filtering commonly observed in neural networks~\cite{rahaman2019spectral}.

The principal innovations of this work address three critical gaps in existing machine learning methods for spatiotemporal reconstruction: 
\begin{enumerate}[nosep]
    \item \textit{Extreme sparsity reconstruction.} The framework achieves effective full-state reconstruction from only 64 sensors covering 1.56\% of the spatial domain---substantially below the density thresholds required by conventional methods. 
    \item \textit{Computational efficiency for operational deployment.} The lightweight architecture enables training and inference on standard hardware within minutes, making it suitable for resource-constrained operational settings where extensive computational infrastructure is unavailable \citep{erichson2020shallow}.
    \item \textit{Hierarchical frequency peeling for heterogeneous fields.} We introduce a novel sequential peeling strategy that decomposes high-frequency corrections into interpretable layers with explicit spectral constraints and frequency exclusion mechanisms. Combined with coordinate-based INR decoders, this approach produces spatially coherent reconstructions that preserve topological structure. 
\end{enumerate}
While demonstrated here on vegetation index, the framework could be generalized beyond Earth observation to broader remote sensing tasks with learnable structure and sparse measurement on heterogeneous spatiotemporal fields.
Code is available at: \url{github.com/xswzaqnjimko/SENDAI_framework}.

\begin{figure}[t]
    \centering
    \includegraphics[width=\columnwidth]{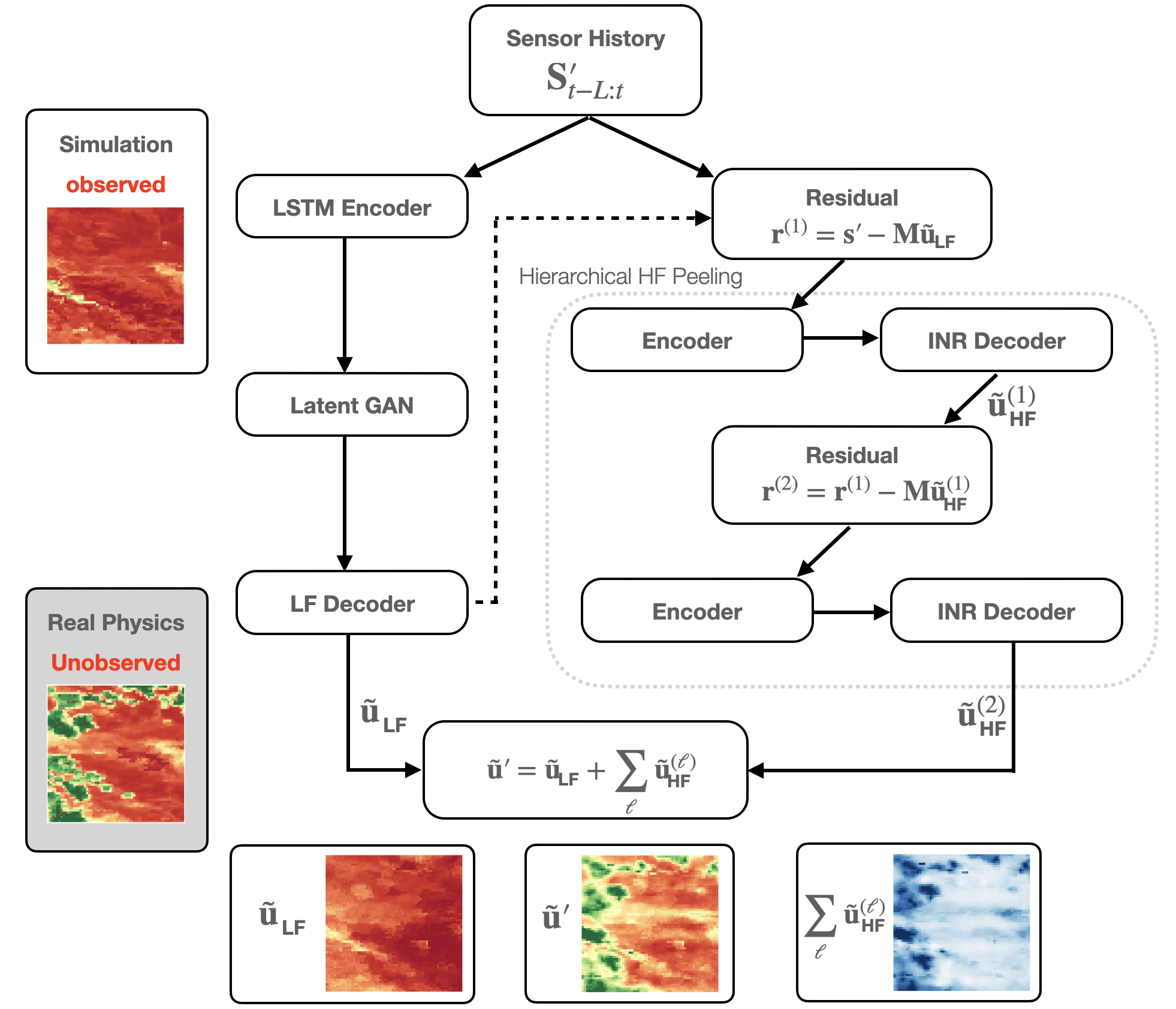}
    \caption{SENDAI architecture. The LF pathway learns dominant dynamics from simulation and aligns to ground truth via a latent GAN. The HF pathway employs sequential peeling layers, each consisting of a sensor residual encoder and a coordinate-based INR decoder, to extract spectrally-distinct corrections.}
    \vspace{-8mm}
    \label{fig:hierarchical_architecture}
\end{figure}

\section{Preliminaries}
\label{sec:data}

We demonstrate SENDAI on satellite remote sensing, reconstructing MODIS-derived NDVI fields across six globally distributed sites spanning Mediterranean, continental, arid, and subtropical climates, and using seasonal periods as proxies for the sim2real domain shift.

\subsection{Remote Sensing Data and Vegetation Index}
\label{sec:modis_ndvi}

We utilize imagery from the Moderate Resolution Imaging Spectroradiometer (MODIS) aboard NASA's Terra and Aqua satellites~\citep{justice2002overview}, which provide continuous global observations with complete coverage every one to two days. Our primary state variable is the Normalized Difference Vegetation Index (NDVI), computed as: 
{
\setlength{\abovedisplayskip}{4pt}
\setlength{\belowdisplayskip}{4pt}
\begin{equation}
    \text{NDVI} = \frac{\rho_{\text{NIR}} - \rho_{\text{Red}}}{\rho_{\text{NIR}} + \rho_{\text{Red}}}
    \label{eq:ndvi}
\end{equation}
}where $\rho_{\text{NIR}}$ and $\rho_{\text{Red}}$ denote surface reflectance in the near-infrared (841--876~nm) and red (620--670~nm) spectral bands, respectively~\citep{tucker1979red, huete2002overview}. NDVI exploits the distinctive spectral signature of photosynthetically active vegetation, ranging from approximately $-0.1$ for water bodies and bare soil to $0.8$--$0.9$ for dense vegetation.  Specifications of MODIS is in Appendix~\ref{app:modis_details}.

\subsection{Data Collection and Experimental Setup}
\label{sec:data_acquisition}

\textbf{Data Acquisition and Processing.} All MODIS imagery was acquired through Google Earth Engine~\citep{gorelick2017google}, merging Terra and Aqua observations to maximize temporal density. For each study site, we define a 15~km $\times$ 15~km region resampled to a standardized $64 \times 64$ pixel grid. Complete specifications are provided in Appendix~\ref{app:data_processing}.

\textbf{Study Sites.} We evaluate our SENDAI framework across six globally distributed study sites spanning diverse climate zones, land cover types, and phenological regimes (Figure~\ref{fig:study_sites_world}). This geographic diversity serves to demonstrate the generalizability of the approach across heterogeneous landscapes and explicitly tests model robustness to out-of-distribution conditions arising from differences in vegetation phenology, soil backgrounds, and atmospheric conditions. Table~\ref{tab:study_sites} summarizes the geographic, climatic, and phenological characteristics of each study site. Details of site biogeochemical characteristics are provided in Appendix~\ref{app:study_sites}.

\textbf{Seasonal Split Strategy and Sensor Configuration.} A fundamental challenge in applying data assimilation frameworks to satellite remote sensing is the absence of a true physics-based simulation model analogous to those available for fluid dynamics or combustion systems. In our formulation, we repurpose the architecture by treating observations from one seasonal period as the ``simulation'' training data and observations from a different season as the ``ground truth'' reality to be reconstructed. The specific seasonal assignments and detailed sparse sensor configuration specifications are provided in Appendix~\ref{app:sensor_config}.

\subsection{Baseline Methods}
\label{sec:baselines}

To contextualize SENDAI's performance, we compare against three established baselines that are widely used in remote sensing and geostatistical workflows for spatiotemporal field reconstruction from sparse observations. All baselines operate under identical constraints: reconstruction from 64 sensors with full temporal context within the ground truth period.
\textbf{\textit{SG+IDW}} applies Savitzky-Golay temporal filtering followed by Inverse Distance Weighting spatial interpolation~\citep{wong2004comparison}, representing standard remote sensing preprocessing. 
\textbf{\textit{HANTS+IDW}} fits harmonic models to capture seasonal phenology~\citep{roerink2000reconstructing} before IDW interpolation, explicitly encoding vegetation seasonality. 
\textbf{\textit{Kriging}} performs Gaussian Process regression with RBF kernel fitted per timestep, evaluated with pretrained hyperparameters from simulation fields. 
Beyond these established geostatistical methods, we additionally evaluate on recent deep learning approach. \textbf{\textit{MMGN}} (Multiplicative and Modulated Gabor Network)~\citep{luo2024continuous} employs an auto-decoder architecture with learnable latent and Gabor filter-based network for continuous field reconstruction from sparse observations. Implementation details for baseline models are provided in Appendix~\ref{app:baselines}.


\section{SENDAI Architecture}
\label{sec:method}

We develop a hierarchical data assimilation architecture that reconstructs full spatial fields from sparse sensor observations, bridging the gap between simplified simulation models and complex multi-physics ground truth systems. The approach builds upon the DA-SHRED framework and Cheap2Rich, its multiscale extension~\citep{bao2025data,bao2026cheap2rich} while introducing two key methodological advances: (i) a \emph{sequential frequency peeling} strategy that decomposes high-frequency corrections into interpretable layers, enabling cleaner spectral separation and improved stability; and (ii) a \emph{coordinate-based implicit neural representation} (INR) for the high-frequency pathway that produces spatially coherent reconstructions. Critically, the full field of the ground truth system is never observed during training---only sparse sensor measurements are available from the target domain. The resulting architecture is lightweight and can be trained on standard hardware without requiring GPU clusters. Figure~\ref{fig:hierarchical_architecture} illustrates the complete pipeline.

\subsection{Problem Formulation}

We adopt the notation established in~\cite{bao2026cheap2rich}. Let $\mathbf{u}_k = \mathbf{u}(\mathbf{x}, t_k) \in \mathbb{R}^n$ denote the full state at time $t_k$, where $n = H \times W$ is the spatial dimension. Sparse sensor measurements are given by $\mathbf{s}_k = \mathbf{M}\mathbf{u}_k \in \mathbb{R}^p$ with $p \ll n$, where $\mathbf{M} \in \mathbb{R}^{p \times n}$ is the sampling operator. The simulation model, governed by $\dot{\mathbf{u}} = \mathcal{N}(\mathbf{u}, \mathbf{x}, t)$, provides full-state training data $\mathbf{X} = [\mathbf{u}_1, \ldots, \mathbf{u}_m]$. For the multi-physics ground truth system, governed by an unknown $\dot{\mathbf{u}}' = \mathcal{M}(\mathbf{u}', \mathbf{x}, t)$, only sensor measurements $\mathbf{S}' = [\mathbf{s}'_1, \ldots, \mathbf{s}'_m]$ are observed.

The reconstruction task seeks a mapping $\mathbf{X}' = F_\theta(\mathbf{S}')$ that estimates the unobserved full state of the ground truth system. Following the multi-scale decomposition~\citep{bao2026cheap2rich,ilersich2025learning}, we write:
{
\setlength{\abovedisplayskip}{4pt}
\setlength{\belowdisplayskip}{4pt}
\begin{equation}
\tilde{\mathbf{u}}'(t) = \tilde{\mathbf{u}}_{\text{LF}}(t) + \tilde{\mathbf{u}}_{\text{HF}}(t),
\label{eq:decomposition}
\end{equation}
}where $\tilde{\mathbf{u}}_{\text{LF}}$ captures dominant dynamics learned from simulation and adapted via latent-space alignment; $\tilde{\mathbf{u}}_{\text{HF}}$ represents fine-scale corrections absent from the simplified model.

\subsection{Low-Frequency (LF) Pathway}
\label{sec:lf_pathway}

The low-frequency pathway follows the DA-SHRED methodology~\citep{bao2025data}, consisting of a temporal encoder trained on simulation data and a latent-space alignment mechanism.

\textbf{Temporal Encoder.} Given sensor time-history $\mathbf{S}'_{t-L:t} = [\mathbf{s}'_{t-L+1}, \ldots, \mathbf{s}'_t] \in \mathbb{R}^{L \times p}$ with $L$ temporal lags, the encoder maps this sequence to a latent representation via a multi-layer LSTM:
{
\setlength{\abovedisplayskip}{4pt}
\setlength{\belowdisplayskip}{4pt}
\begin{equation}
\mathbf{z}_{\text{LF}}(t) = \mathcal{E}_{\text{LF}}(\mathbf{S}'_{t-L:t}; \theta_{\text{enc}}) = \text{LayerNorm}\left(\mathbf{h}^{(K)}_L\right),
\label{eq:lf_encoder}
\end{equation}
}where $\mathbf{h}^{(K)}_L \in \mathbb{R}^{d_z}$ is the final hidden state of the $K$-layer LSTM with hidden dimension $d_z$.

\textbf{Latent-Space GAN Alignment.} To address the distribution mismatch between simulation-derived latents and those induced by ground truth sensor measurements, we employ a residual generator $\mathcal{G}$ that learns to align the latent distributions adversarially~\citep{goodfellow2020generative}:
\begin{equation}
\tilde{\mathbf{z}}_{\text{LF}}(t) = \mathbf{z}_{\text{LF}}(t) + \gamma \cdot \mathcal{G}(\mathbf{z}_{\text{LF}}(t); \theta_{\mathcal{G}}),
\label{eq:gan_alignment}
\end{equation}
where $\gamma$ is a learnable scale parameter initialized small to ensure stable training. The generator and discriminator $\mathcal{D}$ are trained with the standard adversarial objectives as described in~\cite{bao2026cheap2rich}.

\textbf{LF Decoder with Spectral Constraint.} The aligned latent code is decoded and then low-pass filtered to enforce the low-frequency constraint:
\begin{equation}
\tilde{\mathbf{u}}_{\text{LF}}(t) = \mathcal{P}_{k_c}\left(\mathcal{D}_{\text{LF}}(\tilde{\mathbf{z}}_{\text{LF}}(t); \theta_{\text{dec}})\right) \in \mathbb{R}^n,
\label{eq:lf_decoder}
\end{equation}
where $\mathcal{D}_{\text{LF}}$ is a multi-layer MLP with ReLU activations and layer normalization, and $\mathcal{P}_{k_c}$ denotes a low-pass filter that retains only Fourier modes with wavenumber $k \leq k_c$:
\begin{equation}
\mathcal{P}_{k_c}(\mathbf{u}) = \mathcal{F}^{-1}\left(\mathbf{1}_{k \leq k_c} \cdot \mathcal{F}(\mathbf{u})\right),
\label{eq:lowpass}
\end{equation}
where $\mathcal{F}$ and $\mathcal{F}^{-1}$ denote the discrete Fourier transform and its inverse. This explicit spectral constraint ensures a clean separation between LF and HF components.

\subsection{Hierarchical High-Frequency Peeling (HFP)}
\label{sec:hf_peeling}

A key limitation of single-stage high-frequency correction is the difficulty in separating distinct spectral modes that may have different physical origins. We introduce a \emph{hierarchical peeling} structure that sequentially extracts frequency components layer by layer, enabling cleaner spectral separation and improved stability and interpretability.

\textbf{Sequential Residual Computation.} Let $\mathbf{u}^{(0)} = \tilde{\mathbf{u}}_{\text{LF}}$ denote the low-frequency base reconstruction. For $\ell = 1, \ldots, N_{\text{peel}}$ hierarchical layers, we compute:
\begin{align}
\mathbf{r}^{(\ell)}(t) &= \mathbf{s}'(t) - \mathbf{M}\mathbf{u}^{(\ell-1)}(t), \label{eq:residual_computation} \\
\tilde{\mathbf{u}}_{\text{HF}}^{(\ell)}(t) &= \mathcal{H}^{(\ell)}(\mathbf{r}^{(\ell)}(t); \theta_{\text{HF}}^{(\ell)}), \label{eq:hf_layer} \\
\mathbf{u}^{(\ell)}(t) &= \mathbf{u}^{(\ell-1)}(t) + \tilde{\mathbf{u}}_{\text{HF}}^{(\ell)}(t), \label{eq:cumulative}
\end{align}
where $\mathbf{r}^{(\ell)} \in \mathbb{R}^p$ is the sensor residual after all previous corrections, and $\mathcal{H}^{(\ell)}$ is the $\ell$-th high-frequency pathway.

The critical insight is that each layer $\mathcal{H}^{(\ell)}$ sees only the residual \emph{after} all preceding layers have been applied, with gradients detached from previous layers during training. This prevents mode interference and encourages each layer to capture distinct spectral content.

\textbf{Frequency-Guided Sparsity with Exclusion.} To promote interpretable frequency decomposition, each peeling layer is trained with a bandlimited sparsity regularizer that additionally \emph{excludes} frequencies discovered by previous layers. Let $\hat{\mathbf{u}}_{\text{HF}}^{(\ell)}(\mathbf{k})$ denote the 2D Fourier coefficients of the $\ell$-th HF output. The sparsity loss combines three terms:
\begin{equation}
\mathcal{R}_{\text{sparse}}^{(\ell)} = \underbrace{\frac{\|\hat{\mathbf{u}}^{(\ell)}\|_{1,\mathcal{B}}}{\|\hat{\mathbf{u}}^{(\ell)}\|_{2,\mathcal{B}} + \epsilon}}_{\text{in-band L1/L2}} + \beta_1 \mathcal{P}_{\bar{\mathcal{B}}}^{(\ell)} + \beta_2 \mathcal{P}_{\mathcal{E}}^{(\ell)},
\label{eq:sparsity_exclusion}
\end{equation}
where the penalty terms are defined as:
\begin{align}
\mathcal{P}_{\bar{\mathcal{B}}}^{(\ell)} &= \frac{\|\hat{\mathbf{u}}^{(\ell)}\|_{2,\bar{\mathcal{B}}}^2}{\|\hat{\mathbf{u}}^{(\ell)}\|_2^2 + \epsilon} \quad \text{(out-of-band)}, \label{eq:out_of_band} \\
\mathcal{P}_{\mathcal{E}}^{(\ell)} &= \frac{\|\hat{\mathbf{u}}^{(\ell)}\|_{2,\mathcal{E}^{(\ell)}}^2}{\|\hat{\mathbf{u}}^{(\ell)}\|_2^2 + \epsilon} \quad \text{(exclusion)}. \label{eq:exclusion}
\end{align}
Here $\mathcal{B} = \{\mathbf{k} : \|\mathbf{k}\| \leq k_{\max}\}$ is the target frequency band, $\bar{\mathcal{B}}$ its complement, and $\mathcal{E}^{(\ell)} = \bigcup_{j<\ell} \mathcal{E}_j$ is the union of exclusion regions around frequencies discovered by layers $1, \ldots, \ell-1$. The exclusion radius $r_{\text{exc}}$ defining each $\mathcal{E}_j$ creates a buffer zone around discovered frequencies, preventing mode leakage where a subsequent layer partially recaptures the same spectral content with slight offset. The weights $\beta_1, \beta_2 \gg 1$ (typically $\beta_1 = \beta_2 = 100$) strongly penalize energy outside the allowed band and near previously captured modes.

\textbf{Adaptive Top-$k_\ell$ Mode Selection.} For each peeling layer $\ell$, we employ a top-$k_\ell$ sparsity term that encourages energy concentration in $k_\ell$ dominant modes:
\begin{equation}
\mathcal{R}_{\text{topk}}^{(\ell)} = 1 - \frac{\sum_{i=1}^{k_\ell} |\hat{u}_{(i)}|^2}{\|\hat{\mathbf{u}}_{\text{HF}}^{(\ell)}\|_2^2 + \epsilon},
\label{eq:topk_sparsity}
\end{equation}
where $|\hat{u}_{(1)}| \geq \cdots \geq |\hat{u}_{(k_\ell)}|$ are the $k_\ell$ largest Fourier magnitudes. The key challenge is selecting $k_\ell$ adaptively to avoid peeling too much (capturing noise or minor modes) or too little (leaving correlated modes partially captured). We determine $k_\ell$ via spectral analysis of the layer's input residual $\mathbf{r}^{(\ell)}$ prior to training using two criteria:

\textit{(i) Correlation clustering:} Modes with similar frequencies often arise from the same physical phenomenon. We group modes whose frequency radii fall within a bandwidth $\Delta k$:
\begin{equation}
k_\ell^{\text{cluster}} = \left| \left\{ \mathbf{k} : \left| \|\mathbf{k}\| - \|\mathbf{k}^*\| \right| \leq \Delta k \right\} \right|,
\label{eq:k_cluster}
\end{equation}
where $\mathbf{k}^* = \arg\max_{\mathbf{k}} |\hat{r}^{(\ell)}_\mathbf{k}|$ is the dominant frequency in the residual.

\textit{(ii) Energy concentration threshold:} Select $k_\ell$ such that capturing these modes accounts for a target fraction $\rho$ (e.g., $\rho = 0.8$) of the residual's spectral energy:
\begin{equation}
k_\ell^{\text{energy}} = \min \left\{ k : \frac{\sum_{i=1}^{k} |\hat{r}_{(i)}|^2}{\|\hat{\mathbf{r}}^{(\ell)}\|_2^2} \geq \rho \right\}.
\label{eq:k_energy}
\end{equation}

The final selection takes $k_\ell = \max(k_\ell^{\text{cluster}}, k_\ell^{\text{energy}})$, ensuring we capture coherent mode clusters with sufficient energy. For subsequent layers where the residual structure is less pronounced, we may set $k_\ell = \infty$ (equivalently, use bandlimited sparsity only without the top-$k$ constraint), allowing the layer to capture all remaining in-band energy.

\subsection{Coordinate-Based Implicit Neural Representation}
\label{sec:inr}

Direct MLP-based mapping from sparse sensor residuals to full spatial fields often produces ``dotted'' artifacts---localized peaks at sensor locations with poor interpolation elsewhere. This occurs because the network optimizes primarily for sensor-location accuracy without explicit spatial structure.

We address this limitation by replacing the direct MLP decoder with a \emph{coordinate-based implicit neural representation} (INR)~\citep{sitzmann2020implicit,tancik2020fourier} that learns a continuous function over space. The key insight is that querying a shared decoder at \emph{all} spatial coordinates, conditioned on a global latent encoding of the sensor residual, naturally produces smooth interpolation.

\textbf{Architecture.} Each HF peeling layer $\mathcal{H}^{(\ell)}$ consists of two components:

\textit{(i) Sensor Residual Encoder:} Maps the $p$-dimensional sensor residual to a compact latent code:
\begin{equation}
\mathbf{z}_{\text{HF}}^{(\ell)} = \mathcal{E}_{\text{HF}}^{(\ell)}(\mathbf{r}^{(\ell)}; \theta_{\text{enc}}^{(\ell)}) \in \mathbb{R}^{d_{\text{HF}}},
\label{eq:hf_encoder}
\end{equation}
implemented as a multi-layer perceptron with layer normalization.

\textit{(ii) Coordinate-Based Decoder:} For each spatial coordinate $(x, y) \in [0, 1]^2$ (normalized), the decoder computes:
\begin{equation}
u_{\text{HF}}^{(\ell)}(x, y) = \gamma^{(\ell)} \cdot \mathcal{D}_{\text{INR}}^{(\ell)}\bigl([\text{PE}(x, y); \mathbf{z}_{\text{HF}}^{(\ell)}]\bigr),
\label{eq:inr_decoder}
\end{equation}
where $\text{PE}(\cdot)$ is a Fourier positional encoding, $[\cdot\,; \cdot]$ denotes concatenation, and $\gamma^{(\ell)}$ is a learnable scale parameter.

\textbf{Fourier Positional Encoding.} Following~\cite{tancik2020fourier}, we encode spatial coordinates using sinusoidal features at $L$ frequency bands $\{\sigma_j\}_{j=1}^L$, typically log-spaced from $1$ to $\sigma_{\max}$:
\begin{align}
\text{PE}(x, y) = \bigl[x, y,\, &\sin(2\pi \sigma_1 x), \cos(2\pi \sigma_1 x), \ldots, \nonumber \\
&\sin(2\pi \sigma_L y), \cos(2\pi \sigma_L y)\bigr].
\label{eq:fourier_pe}
\end{align}
This encoding enables the MLP to learn high-frequency spatial patterns that would otherwise be difficult due to the spectral bias of neural networks toward low frequencies~\citep{rahaman2019spectral}.

\textbf{Spatial Smoothness Regularization.} To further encourage spatially coherent outputs, we incorporate a regularizer on the spatial structure of the HF field. The choice of regularizer should be informed by prior knowledge of the expected HF characteristics:

For systems where the high-frequency correction is expected to exhibit \emph{smooth} spatial structure (e.g., gentle gradients or diffusive processes), we employ a Laplacian regularizer that penalizes curvature:
\begin{equation}
\mathcal{R}_{\text{lap}}^{(\ell)} = \frac{1}{|\Omega|} \sum_{(i,j) \in \Omega} \left|\nabla^2 u_{\text{HF}}^{(\ell)}(i, j)\right|^2,
\label{eq:laplacian_reg}
\end{equation}
where $\nabla^2 u = u_{i+1,j} + u_{i-1,j} + u_{i,j+1} + u_{i,j-1} - 4u_{i,j}$ is the discrete Laplacian. This allows sharp but smooth features while suppressing spurious high-frequency oscillations.

Conversely, for systems where the HF correction contains \emph{sharp discontinuities} or localized features (e.g., shock fronts, edges, or concentrated anomalies), the Laplacian regularizer would inappropriately penalize physically meaningful curvature. In such cases, a gradient-based total variation regularizer $\mathcal{R}_{\text{TV}}^{(\ell)} = \sum_{i,j} |\nabla u_{\text{HF}}^{(\ell)}(i,j)|$ preserves edges while promoting piecewise smoothness, or an edge-preserving bilateral formulation using Huber loss may be employed. The appropriate regularizer should be selected based on domain knowledge of the underlying physics; we provide detailed formulations in Appendix~\ref{app:smoothness}.

\textbf{Full HF Layer Loss.} The training objective for layer $\ell$ is:
\begin{align}
\mathcal{L}^{(\ell)} = \mathcal{L}_{\text{sensor}}^{(\ell)} &+ \lambda_{\text{sp}} \mathcal{R}_{\text{sparse}}^{(\ell)} + \lambda_{\text{topk}} \mathcal{R}_{\text{topk}}^{(\ell)} \nonumber \\
&+ \lambda_{\text{sm}} \mathcal{R}_{\text{smooth}}^{(\ell)} + \lambda_{\text{mag}} \mathcal{L}_{\text{mag}}^{(\ell)},
\label{eq:hf_loss}
\end{align}
where $\mathcal{L}_{\text{sensor}}^{(\ell)} = \|\mathbf{M}\tilde{\mathbf{u}}_{\text{HF}}^{(\ell)} - \mathbf{r}^{(\ell)}\|_2^2$ is the sensor matching loss, $\mathcal{R}_{\text{smooth}}^{(\ell)}$ is the spatial regularizer, and $\mathcal{L}_{\text{mag}}^{(\ell)} = [\max(0, \|\tilde{\mathbf{u}}_{\text{HF}}^{(\ell)}\|_1 - \tau)]^2$ is a magnitude constraint preventing the HF component from dominating. When $k_\ell = \infty$, we set $\lambda_{\text{topk}} = 0$.

\subsection{Training Pipeline}
\label{sec:training}

The complete training procedure proceeds in three stages. \textbf{\textit{Stage 1}} (SHRED on Simulation): Train the base SHRED model (LSTM encoder + decoder) on simulation data with either sparse sensor histories or full-state supervision to learn the dominant dynamics and establish a smooth spatial decoder.
\textbf{\textit{Stage 2}} (Latent Alignment): Train the latent transformation $\mathcal{G}$ and discriminator $\mathcal{D}$ via adversarial learning to align simulation and ground truth latent distributions. 
\textbf{\textit{Stage 3}} (Hierarchical HF Peeling): For each layer $\ell = 1, \ldots, N_{\text{peel}}$, freeze all parameters except $\theta_{\text{HF}}^{(\ell)}$, determine $k_\ell$ adaptively via spectral analysis (Eqs.~\ref{eq:k_cluster}--\ref{eq:k_energy}), train with a warmup schedule where $\lambda_{\text{sp}} = 0$ for the first $E_{\text{warm}}$ epochs before ramping linearly to the target value, identify dominant frequencies after training and add them to the exclusion set $\mathcal{E}^{(\ell)}$ for subsequent layers, then fine-tune with reduced sparsity weight $\lambda'_{\text{sp}} = 0.1 \lambda_{\text{sp}}$.

\textbf{Inference}
Given sensor history $\mathbf{S}'_{t-L:t}$ and current measurements $\mathbf{s}'(t)$, inference proceeds as follows. The LF pathway computes $\mathbf{z}_{\text{LF}} = \mathcal{E}_{\text{LF}}(\mathbf{S}'_{t-L:t})$, applies the learned alignment $\tilde{\mathbf{z}}_{\text{LF}} = \mathbf{z}_{\text{LF}} + \gamma \cdot \mathcal{G}(\mathbf{z}_{\text{LF}})$, and decodes $\mathbf{u}^{(0)} = \mathcal{D}_{\text{LF}}(\tilde{\mathbf{z}}_{\text{LF}})$. The HF pathway then iteratively refines the reconstruction: for $\ell = 1, \ldots, N_{\text{peel}}$, compute residual $\mathbf{r}^{(\ell)} = \mathbf{s}'(t) - \mathbf{M}\mathbf{u}^{(\ell-1)}$ and update $\mathbf{u}^{(\ell)} = \mathbf{u}^{(\ell-1)} + \mathcal{H}^{(\ell)}(\mathbf{r}^{(\ell)})$. The final reconstruction is $\tilde{\mathbf{u}}'(t) = \mathbf{u}^{(N_{\text{peel}})}$.




\section{Results}
\label{sec:results}

We evaluate the SENDAI framework across six geographically diverse study sites, assessing reconstruction of heterogeneous spatiotemporal NDVI fields from sparse sensor measurements. The evaluation encompasses two architectural variants: (i) the simplified SENDAI Jr. pipeline for sites with primarily low-frequency discrepancies, and (ii) the full SENDAI hierarchical multiscale architecture for sites requiring high-frequency correction. All experiments employ 64 sensors ($\sim$1.5\% of pixels). We adopt the Structural Similarity Index Measure (SSIM) as our primary performance metric, as it captures the preservation of spatial patterns and topological structure that RMSE alone cannot assess---a smoothed reconstruction may achieve reasonable RMSE while completely destroying field boundaries and land cover discontinuities.

\subsection{Synthetic Analysis}
\label{sec:synthetic}

Before evaluating full 2D NDVI reconstruction, we validate the HFP methodology on controlled systems where ground truth decomposition is known analytically. We present two complementary experiments: (i) a synthetic traveling wave system with analytically specified frequencies, and (ii) a 1D slice extracted from real MODIS NDVI data where simulation and ground truth exhibits clear spectral discrepancy.

\begin{figure}[t]
    \centering
    \includegraphics[width=\columnwidth]{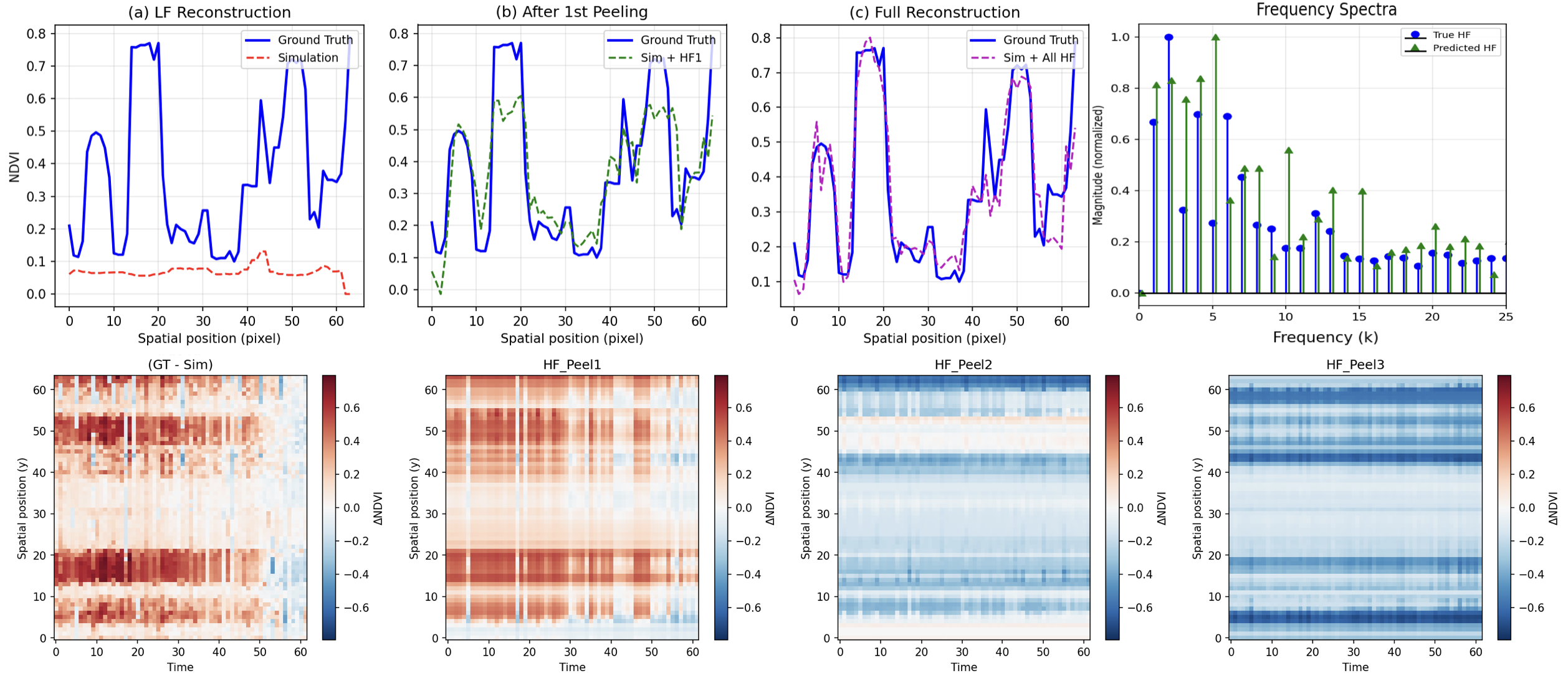}
    \caption{Hierarchical frequency peeling on a 1D NDVI slice from the Tarim Basin site. Top row presents single time-point reconstruction; bottom row presents spatiotemporal HF decomposition.}
    \vspace{-4mm}
    \label{fig:synthetic_comparison2}
\end{figure}

\textbf{Traveling Wave System.} We construct a spatiotemporal field composed of three distinct traveling waves.
{
\setlength{\abovedisplayskip}{4pt}
\setlength{\belowdisplayskip}{4pt}
\begin{align}
u(x,t) = \sum_{n=1}^{3} A_n \sin\!\left(k_n x - \omega_n t\right)
\end{align}
}Figure~\ref{fig:synthetic_comparison} compares two HF correction strategies: joint discovery versus hierarchical peeling. Both achieve comparable reconstruction accuracy in terms of RMSE, but the joint approach produces a relatively noisy frequency spectrum with energy spread across multiple modes beyond the targets (panel a, rightmost), whereas hierarchical peeling yields clean outputs for each mode (panel b, rightmost). This spectral purity translates to improved fine-scale reconstruction, particularly in regions where the two modes interfere constructively or destructively. The decomposition also enables interpretable analysis of individual frequency contributions—critical for applications where different spectral components have distinct physical origins. Additionally, the hierarchical strategy also exhibits more stable training dynamics and supports modular refinement. Full experimental details and extended analysis is provided in Appendix~\ref{app:synthetic}.

\textbf{NDVI 1D Validation.} To validate hierarchical peeling on remote sensing data with inherent noise, outliers, and non-stationary dynamics, we extract a 1D slice from the Tarim Basin site (Figure~\ref{fig:ndvi_slice}). 

Despite the HF residual exhibits widespread energy, hierarchical peeling successfully captures the majority of the modes (Figure~\ref{fig:synthetic_comparison2}) while producing interpretable components that correspond to distinct spatialtemporal scales of the landscape. The first layer (HF$_1$) captures coherent spatio-temporal variability consistent with climate-driven phenological offsets (temperature/episodic weather forcing) modulated by elevation, while the second layer (HF$_2$) isolates a largely time-invariant spatial gradient suggestive of persistent hydrological controls (e.g. water availability) that differ between spring simulations and summer--autumn observations. A third temporally stable component (HF$_3$) plausibly reflects edaphic heterogeneity (e.g. texture, salinity, nutrient level) that the low-frequency model cannot resolve. 

Success on data without clean sinusoidal patterns demonstrates that hierarchical peeling is robust to realistic data imperfections. Extended analysis is provided in Appendix~\ref{app:synthetic}.


\subsection{Baseline Methods Performance}
\label{sec:results_baselines}

\begin{figure}[t]
\centering
\includegraphics[width=\columnwidth]{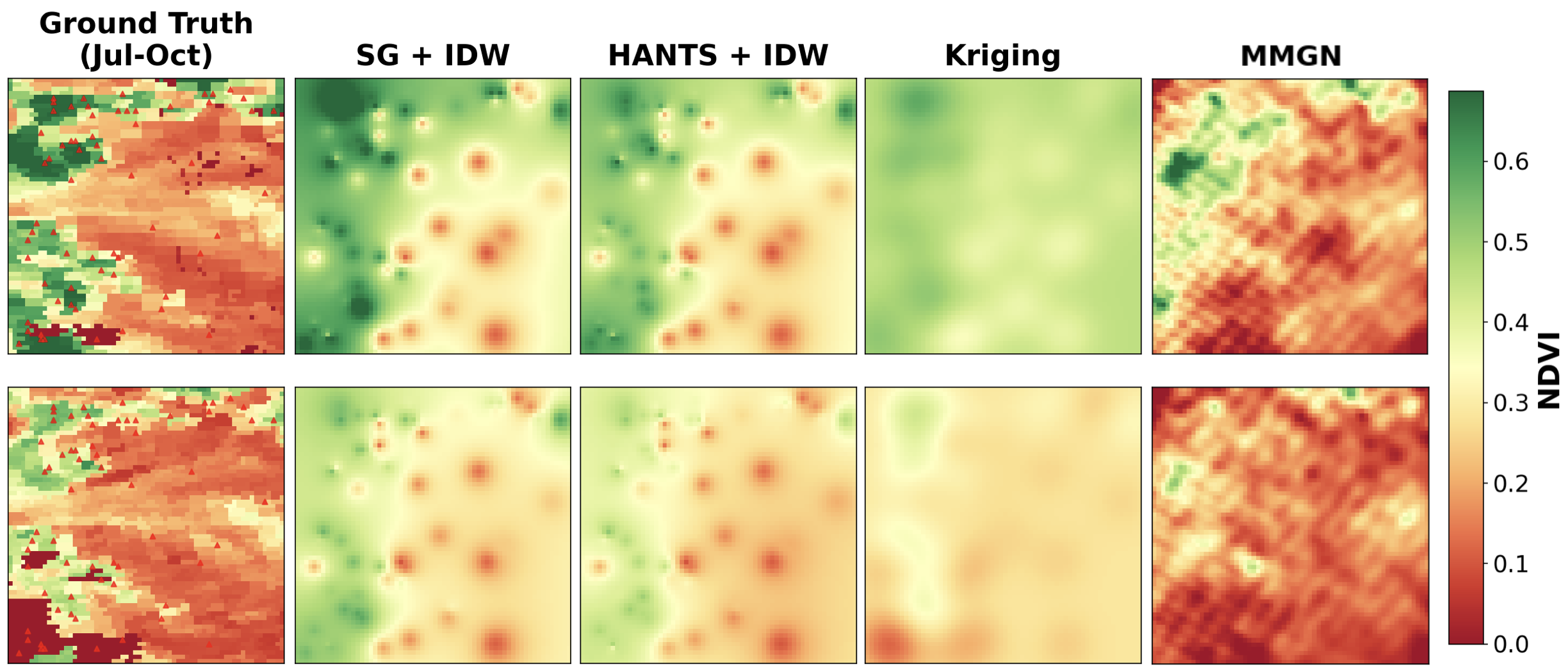}
\caption{Baseline reconstruction comparison for the Tarim Basin site. Red markers indicate sensor locations.}
\vspace{-4mm}
\label{fig:baseline_example}
\end{figure}

We establish baseline performance using three geostatistical methods for spatiotemporal reconstruction---Savitzky-Golay filtering with IDW (SG+IDW), Harmonic Analysis of Time Series with IDW (HANTS+IDW), and Kriging---as well as MMGN, a recent implicit neural representation approach. Figure~\ref{fig:baseline_example} illustrates baseline reconstruction quality on the Tarim Basin site, which presents challenging heterogeneous spatial structure due to sharp mountain-basin boundaries. A critical observation emerges: \emph{baseline methods fundamentally fail to preserve the topological structure of spatial patterns despite achieving moderate RMSE values}. These deficiencies underscore the importance of SSIM (structural similarity index measure) for evaluating structural preservation. MMGN, despite its strong performance on climate and oceanographic datasets, exhibits comparable limitations under our extreme sparsity regime (64 sensors, 1.56\% coverage) and heterogeneous fine-scale landscapes.

\subsection{SENDAI Jr. Reconstruction Performance}
\label{sec:results_dashred}

For sites where seasonal domain shift manifests predominantly as low-frequency distributional changes, the two-stage alignment alone from SENDAI (SENDAI Jr.) provides effective reconstruction. Implementation details are in Appendix~\ref{app:simplified_details}. Table~\ref{tab:dashred_results} summarizes the performance.

SENDAI Jr. achieves substantial SSIM improvements over all baselines: +120\% for Central Valley, +185\% for Corn Belt, and +98\% for Guadalquivir Valley. These improvements demonstrate that SENDAI Jr. successfully preserves spatial topology---field boundaries, vegetation gradients, and land cover patterns---that interpolation-based methods systematically destroy. RMSE improvements confirm quantitative accuracy alongside structural fidelity.

\begin{figure}[t]
\centering
\includegraphics[width=\columnwidth]{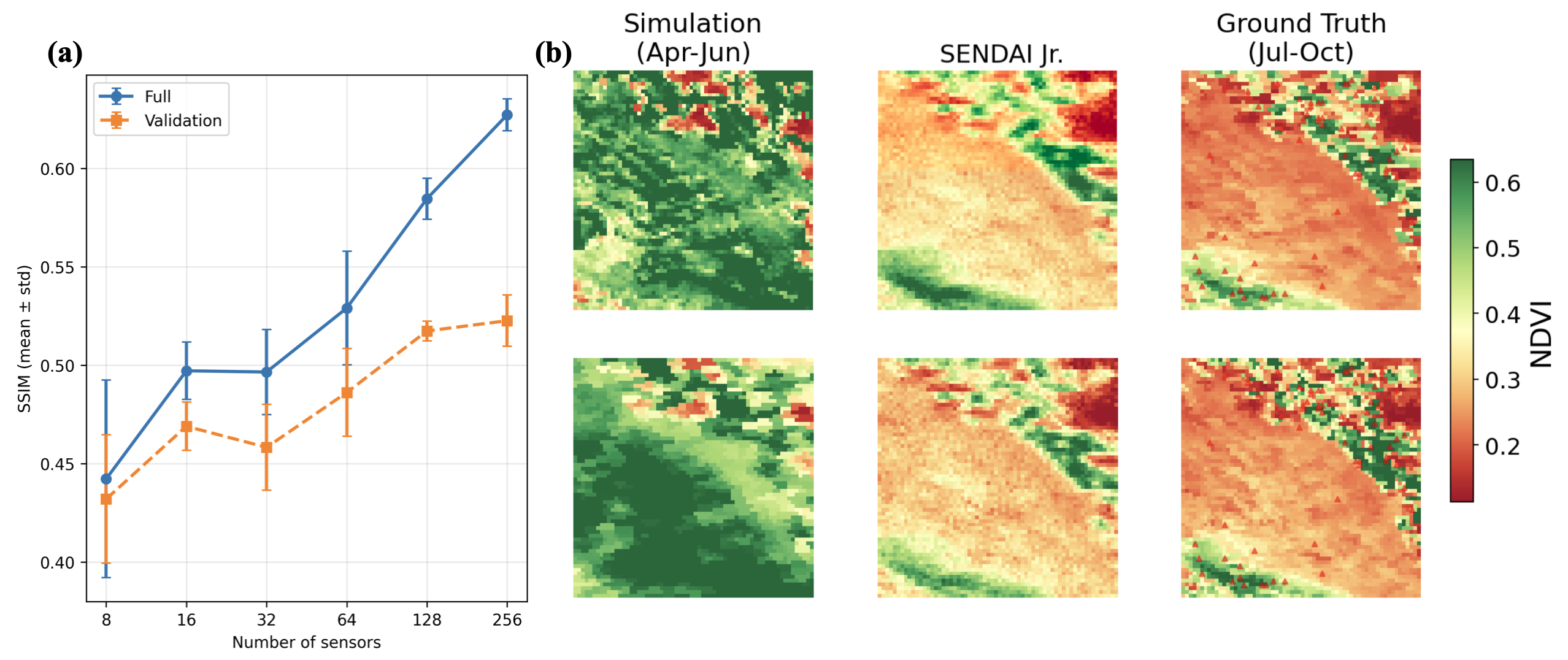}
\caption{(a) Sensor sensitivity results (from five independent runs) on the Tarim Basin site for the full SENDAI framework. (b) SENDAI Jr. reconstruction for the Central Valley site.}
\label{fig:dashred_cv_main}
\end{figure}

Figure~\ref{fig:dashred_cv_main}b presents reconstruction for the Central Valley site. SENDAI Jr. preserves the essential topological structure including the spatial arrangement of high- and low-NDVI regions. Site-specific details are provided in Appendix~\ref{app:dashred_details}.

\begin{table}[t]
\centering
\caption{SENDAI Jr. reconstruction performance compared with baselines. The first row quantifies the phenological domain shift. Relative improvement from best baseline to SENDAI Jr. is indicated. Best results per site are \textbf{bolded}. All of the experiments below are averaged from five independent runs. }
\label{tab:dashred_results}
\resizebox{\columnwidth}{!}{%
\begin{tabular}{llccc}
\toprule
& & \textbf{Central Valley} & \textbf{Corn Belt} & \textbf{Guadalquivir} \\
\midrule
Sim.\ vs.\ GT & RMSE & 0.1965 & 0.4501 & 0.2387 \\
 & SSIM & 0.4751 & 0.0312 & 0.2464 \\
\midrule
SG + IDW & RMSE & 0.1447 & 0.1712 & \textbf{0.1444} \\
 & SSIM & 0.2612 & 0.1588 & 0.1849 \\
\midrule
HANTS + IDW & RMSE & 0.1451 & 0.1866 & 0.1496 \\
 & SSIM & 0.2504 & 0.1498 & 0.1755 \\
\midrule
Kriging & RMSE & 0.1634 & 0.1596 & 0.1481 \\
 & SSIM & 0.0922 & 0.0312 & 0.0878 \\
\midrule
\textbf{SENDAI Jr.} & RMSE & \textbf{0.1068} & \textbf{0.1103} & 0.1474 \\
 & SSIM & \textbf{0.5747} & \textbf{0.4530} & \textbf{0.3655} \\
\midrule
\rowcolor{red!20}\multicolumn{2}{l}{Improvement} & +120.0\% & +185.3\% & +97.7\% \\
\bottomrule
\end{tabular}%
}
\end{table}

\subsection{Full SENDAI Performance}
\label{sec:results_multiscale}

For sites exhibiting complex phenological dynamics and pronounced spatial heterogeneity, we deploy SENDAI with full hierarchical multiscale architecture with multiple peeling layers. Table~\ref{tab:multiscale_results} presents detailed result comparison.

The full SENDAI hierarchical architecture achieves the highest SSIM across all three sites: 0.4668 (Imperial Valley), 0.4777 (Tarim Basin), and 0.3354 (Riverina). These represent substantial improvements over both baselines and Cheap2Rich, with SSIM gains of 15.5\% (Imperial Valley), 36.3\% (Tarim Basin), and 21.5\% (Riverina) from Cheap2Rich to the full SENDAI pipeline. The Tarim Basin site exhibits the most dramatic improvement, where sharp mountain-basin boundaries require high-frequency corrections that smooth decoders cannot capture.

\begin{figure}[t]
\centering
\includegraphics[width=\columnwidth]{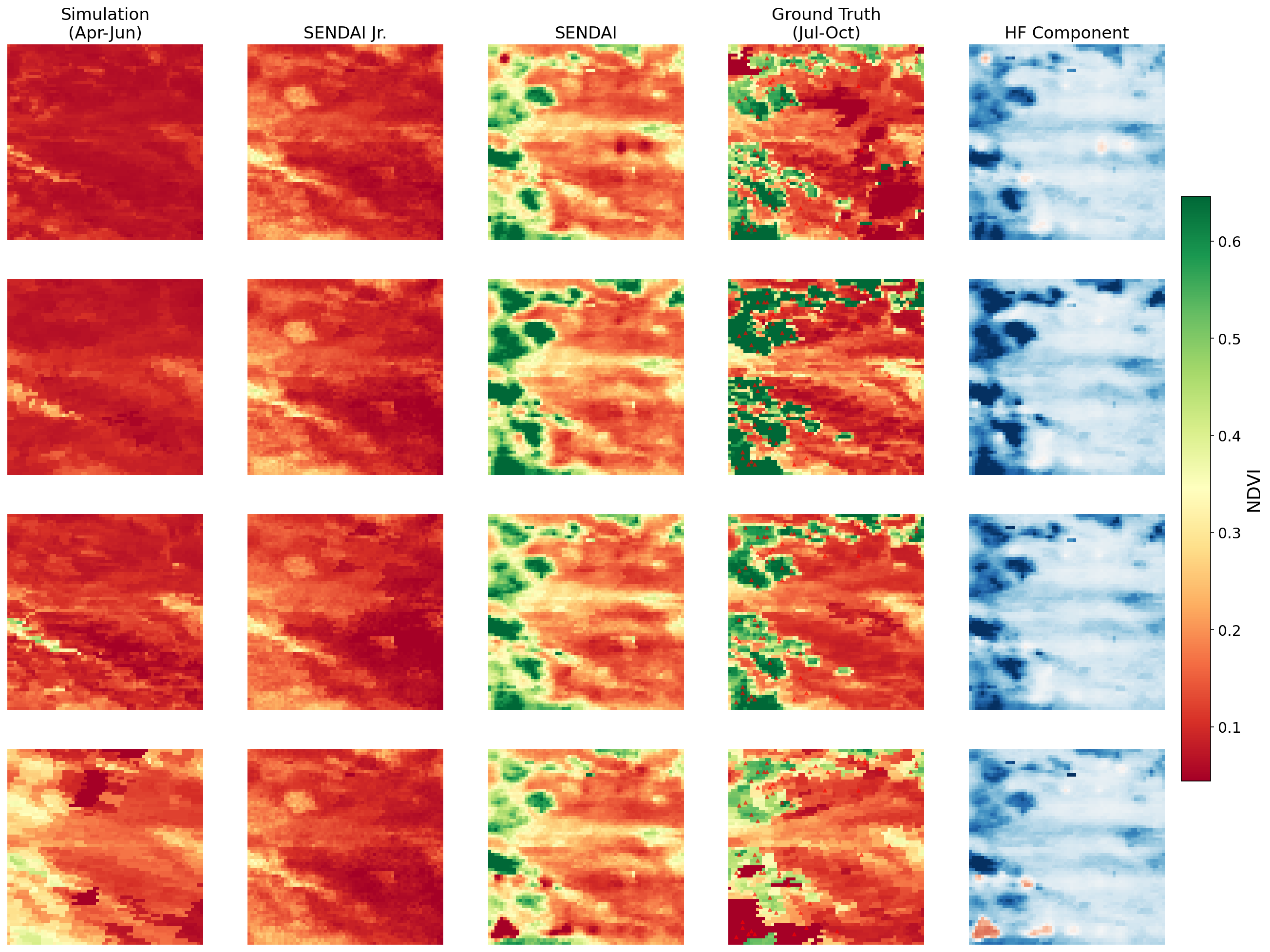}
\caption{Full SENDAI hierarchical multiscale DA-SHRED reconstruction for the Tarim Basin site.}
\vspace{-8mm}
\label{fig:multiscale_xj_main}
\end{figure}

Figure~\ref{fig:multiscale_xj_main} presents the full hierarchical reconstruction result for the Tarim Basin site. While all baseline methods fail to resolve sharp boundaries, SENDAI successfully reconstructs these landscape features. The learned HF component exhibits coherent spatial structure aligned with boundaries, confirming that peeling layers discover physically meaningful corrections. Sensitivity analysis on sensor number is provided in Figure~\ref{fig:dashred_cv_main}a and in Appendix~\ref{appendix:sensor_sensitivity}.

Qualitative examination (Figures~\ref{fig:baseline_iv_app}--\ref{fig:multiscale_aus_app} in Appendix~\ref{app:multiscale_details}) reveals consistent pattern reconstructions across sites, compared with baselines. The SENDAI Jr. base reconstruction recovers mesoscale spatial patterns, while the learned HF correction exhibits site-dependent characteristics: coherent structure correlated with field boundaries for Imperial Valley, sharp landscape boundaries for Tarim Basin, and more diffusive structure for Riverina consistent with gradual spatial transitions. Reconstruction quality remains stable across temporal frames, indicating generalizable representations. Site-specific details are provided in Appendix~\ref{app:multiscale_details}.

\subsection{Computational Efficiency}
\label{sec:results_efficiency}

SENDAI is remarkably efficient computationally. Unlike contemporary deep learning approaches requiring GPU clusters~\citep{meraner2020cloud, cresson2018framework}, SENDAI operates on standard CPU hardware with training times measured in minutes per site. This efficiency, deriving from the shallow architecture and exploitation of Takens' embedding theorem, enables full-state reconstruction from low-dimensional sensor histories without requiring dense spatial supervision. Detailed operational scenarios and quantitative analysis are provided in Appendix~\ref{app:hardware_efficiency_details}.

\begin{table}[t]
\centering
\caption{SENDAI reconstruction performance compared with baselines. Relative improvements from Cheap2Rich are indicated. Best results per site are \textbf{bolded}. All of the experiments below are averaged from five independent runs.}
\label{tab:multiscale_results}
\resizebox{\columnwidth}{!}{%
\begin{tabular}{llccc}
\toprule
& & \textbf{Imperial Valley} & \textbf{Tarim Basin} & \textbf{Riverina} \\
\midrule
Sim.\ vs.\ GT & RMSE & 0.2157 & 0.2077 & 0.2778 \\
 & SSIM & 0.3488 & 0.3448 & 0.1115 \\
\midrule
SG + IDW & RMSE & 0.1599 & 0.1794 & 0.1603 \\
 & SSIM & 0.1123 & 0.1308 & 0.1359 \\
\midrule
HANTS + IDW & RMSE & 0.1603 & 0.1806 & 0.1669 \\
 & SSIM & 0.1049 & 0.1214 & 0.1245 \\
\midrule
Kriging & RMSE & 0.1591 & 0.2163 & \textbf{0.1495} \\
 & SSIM & 0.0916 & 0.0449 & 0.0272 \\
\midrule
MMGN & RMSE & 0.1786 & 0.1783 & 0.3233 \\
 & SSIM & 0.0778 & 0.1226 & 0.0798 \\
\midrule
Cheap2Rich & RMSE & 0.1708 & 0.1827 & 0.1537 \\
 & SSIM & 0.4041 & 0.3505 & 0.2761 \\
\midrule
Cheap2Rich+HFP & RMSE & 0.1588 & 0.1466 & 0.2526 \\
 & SSIM & 0.4411 & 0.4257 & 0.3158 \\
\midrule
\textbf{SENDAI} & RMSE & \textbf{0.1486} & \textbf{0.1208} & 0.1823 \\
 & SSIM & \textbf{0.4668} & \textbf{0.4777} & \textbf{0.3354} \\
 \midrule
\rowcolor{red!20} Improvement & HFP & +9.2\% & +21.5\% & +14.4\% \\
\rowcolor{red!20} from Cheap2Rich & HFP+INR & +15.5\% & +36.3\% & +21.5\% \\
\bottomrule
\end{tabular}%
}
\end{table}

\section{Conclusions and Future Directions}
\label{sec:conclusions}


The experimental results show that SENDAI robustly reconstructs heterogeneous spatiotemporal NDVI fields across diverse geographic and climatic settings, adapting from simulation to ground-truth observations across seasonal boundaries while recovering mesoscale patterns and fine-scale structure via hierarchical peeling. Performance depends on landscape complexity: SENDAI Jr. suffices for smoother fields with persistent spatial structure, whereas sites with sharp boundaries and sub-seasonal dynamics benefit from the full SENDAI framework with hierarchical structure.


The methodological contributions extend beyond NDVI reconstruction. SENDAI provides a general template for sparse-observation reconstruction of heterogeneous spatiotemporal fields: establish spatial priors from data-rich reference sources, adapt to sparse-measurement targets via latent-space alignment, and employ hierarchical frequency peeling when fine-scale corrections are needed. This template is applicable across Earth observation tasks such as soil moisture~\citep{mohanty2017soil}, land surface temperature~\citep{li2013satellite}, snow dynamics~\citep{lievens2019snow}, and flood mapping~\citep{schumann2015microwave}, and may extend to bandwidth-constrained settings (e.g., deep-space monitoring~\citep{de2011reliability}) wherever coherent fields, reference data, and sparse target observations are available. Additional discussion is provided in Appendix~\ref{app:domain_adaptation_details}.


Several limitations and future directions remain. First, SENDAI assumes stationary spatial structure between simulation and ground truth periods, which may break under land cover change, management shifts, or disturbances; extending the framework to non-stationary settings is an important next step. Second, our current random sensor placement could be improved through information-theoretic design~\citep{bao2025information, santos2023development}, potentially reducing sensor requirements. Third, although we observed transferability across regions, rigorous cross-continent generalization tests are still needed to distinguish universal from region-specific representations. Finally, extending to multivariate reconstruction (e.g., NDVI, land surface temperature, soil moisture) could leverage cross-variable correlations via the shared latent space.

\section*{Impact Statement}

In this paper we investigate the applications of Machine Learning to the frontier of remote sensing. The SENDAI framework enables accurate heterogeneous spatiotemporal reconstruction and domain adaptation from severely sparse sensor observations, with potential benefits for environmental monitoring, space exploration, disaster response, and climate analysis, particularly in resource-constrained areas where dense observations are expensive or unavailable.




\section*{Acknowledgments}

This work was supported in part by the US National Science Foundation (NSF) AI Institute for Dynamical Systems (dynamicsai.org), grant 2112085. JNK further acknowledges support from the Air Force Office of Scientific Research  (FA9550-24-1-0141).

\bibliography{refs}
\bibliographystyle{icml2026}

\newpage
\appendix
\onecolumn

\section{MODIS Platform Details}
\label{app:modis_details}

The Moderate Resolution Imaging Spectroradiometer (MODIS) is a key instrument aboard NASA's Terra and Aqua satellites, launched in 1999 and 2002 respectively, providing continuous global observations for over two decades~\citep{justice2002overview}. MODIS acquires data in 36 spectral bands ranging from 0.405 to 14.385~$\mu$m, with spatial resolutions of 250~m, 500~m, and 1000~m depending on the band. The instrument's 2330~km viewing swath width enables complete global coverage every one to two days, making it uniquely suited for monitoring dynamic land surface processes at regional to continental scales.

The Terra satellite follows a descending sun-synchronous orbit with a 10:30 AM local equatorial crossing time, while Aqua follows an ascending orbit with a 1:30 PM crossing. This complementary configuration provides up to two observations per day for any given location, significantly increasing the probability of obtaining cloud-free imagery. In this study, we utilize the Collection 6.1 daily surface reflectance products (MOD09GA from Terra and MYD09GA from Aqua), which provide atmospherically corrected surface reflectance values processed using the Second Simulation of the Satellite Signal in the Solar Spectrum (6S) radiative transfer model.

The Normalized Difference Vegetation Index (NDVI) exploits the distinctive spectral signature of photosynthetically active vegetation, which strongly absorbs red light for photosynthesis while reflecting near-infrared (NIR) radiation due to leaf cellular structure~\citep{tucker1979red}. NDVI values typically range from approximately $-0.1$ for water bodies and bare soil to $0.8$--$0.9$ for dense, healthy vegetation, responding sensitively to changes in chlorophyll content, leaf area, and vegetation fraction~\citep{huete2002overview}.

\section{Data Processing Pipeline Details}
\label{app:data_processing}

All MODIS imagery was acquired through Google Earth Engine (GEE)~\citep{gorelick2017google}, a cloud-based platform providing direct access to petabyte-scale geospatial archives. For each study site, imagery from both Terra and Aqua sensors was merged to maximize temporal density, yielding up to two potential observations per day.

Cloud contamination is addressed using the \texttt{state\_1km} quality assurance band, retaining only pixels flagged as ``clear'' (bit pattern 00 in bits 0--1). Images with less than 70\% valid pixel coverage are excluded to ensure spatial coherence. For each study site, we define a 15~km $\times$ 15~km square region centered on representative coordinates, resampled to a standardized $64 \times 64$ pixel grid at fine resolution. This fixed grid dimension facilitates consistent architecture across study sites and enables direct comparison of model performance.

To ensure consistent temporal density while accounting for variable cloud cover, we employ an equally-spaced sampling strategy targeting 70--90 valid images per approximately 90-day observation period. In practice, persistent cloud cover at several sites (particularly tropical and monsoon-affected regions) limited acquisition to 40--50 valid images per period, testing the framework's robustness under reduced temporal sampling.

For each study site, the data generation pipeline produces simulation data (from one seasonal period) and ground truth data (from a different seasonal period). The simulation data serves to train the base SHRED architecture, learning a latent representation of the state space structure. The ground truth data provides sensor-only observations for the SENDAI, where the model learns to adapt its latent space to the distributional shift between seasons while reconstructing full-state fields from sparse measurements.

\section{Study Site Descriptions}
\label{app:study_sites}

\begin{figure*}[t]
    \centering
    \includegraphics[width=\textwidth]{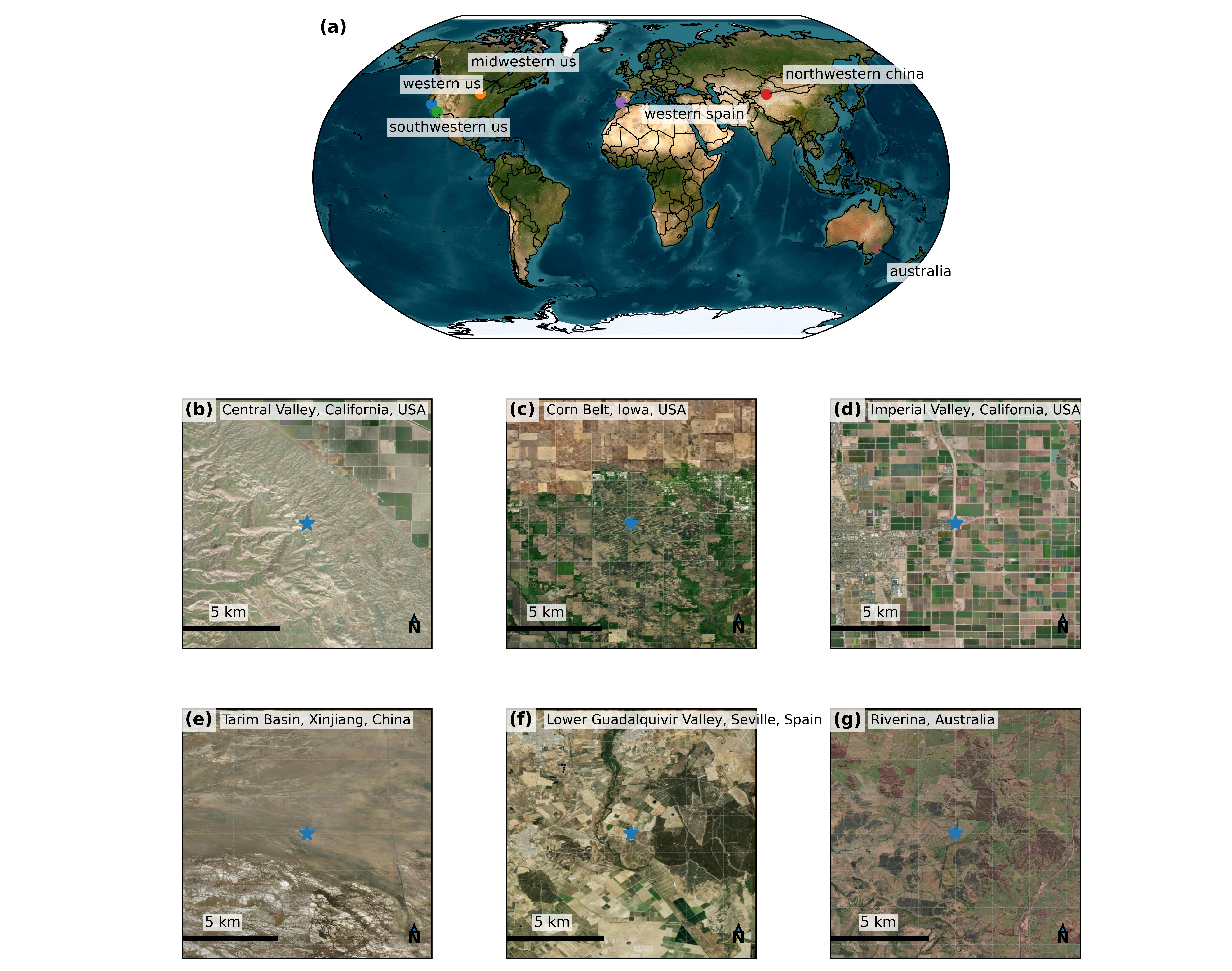}
    \caption{Study sites for NDVI reconstruction experiments. (a) Global distribution of the eight study areas spanning North America, South America, Europe, Asia, and Australia. (b--h) Local 32~km $\times$ 32~km windows centered at each site; stars denote site centers. The basemap imagery shown is a rolling mosaic for geographic context only. Administrative boundaries are from Natural Earth; imagery tiles are from Esri World Imagery or Sentinel-2 cloudless composites.}
    \label{fig:study_sites}
\end{figure*}

\begin{table*}[t]
\centering
\caption{Study site characteristics and observation periods. Sites are grouped by the architecture variant employed: SENDAI Jr. (simpler seasonal transitions) or SENDAI (complex phenological dynamics requiring high-frequency correction). All data are from 2023. Climate classifications follow the K\"oppen system: Csa = hot-summer Mediterranean (dry summers, wetter winters); Dfa = hot-summer humid continental (no dry season, cold winters); BWh = hot desert (extremely arid, very hot); BWk = cold desert (arid but cooler, with colder winters); Cfa = humid subtropical (no dry season, hot humid summers, mild winters).}
\label{tab:study_sites}
\vspace{0.5em}
\small
\resizebox{\columnwidth}{!}{%
\begin{tabular}{llcccccc}
\toprule
\textbf{Region ID} & \textbf{Location Name} & \textbf{Center} & \textbf{Climate} & \textbf{Land Cover} & \textbf{Sim.} & \textbf{GT} & \textbf{Model} \\
 & & \textbf{(Lon, Lat)} & & & & & \\
\midrule
\multicolumn{8}{l}{\textit{North America}} \\
western\_us & Central Valley, CA & $(-120.5, 36.5)$ & Csa & Irrigated cropland & Apr--Jun & Jul--Oct & SENDAI Jr. \\
midwestern\_us & Corn Belt, IA & $(-93.5, 42.0)$ & Dfa & Rainfed cropland & Apr--Jun & Jul--Oct &  SENDAI Jr. \\
southwestern\_us & Imperial Valley, CA & $(-115.5, 32.8)$ & BWh & Irrigated cropland & Apr--Jun & Jul--Oct &  SENDAI \\
\midrule
\multicolumn{8}{l}{\textit{Europe}} \\
western\_spain & Guadalquivir Valley & $(-6.25, 37.25)$ & Csa & Mixed agriculture & Feb--Apr & Sep--Dec &  SENDAI Jr. \\
\midrule
\multicolumn{8}{l}{\textit{Asia}} \\
northwestern\_china & Tarim Basin & $(83.5, 41.5)$ & BWk & Oasis agriculture & Apr--Jun & Jul--Oct &  SENDAI \\
\midrule
\multicolumn{8}{l}{\textit{Southern Hemisphere}} \\
southeasthern\_australia & Riverina & $(147.5, -35.5)$ & Cfa & Mixed cropping & Feb--Apr & Sep--Dec &  SENDAI \\
\bottomrule
\end{tabular}%
}
\end{table*}

\subsection{Western United States: Central Valley, California}

The Central Valley represents one of the world's most productive agricultural regions, characterized by Mediterranean climate (K\"oppen Csa) with hot, dry summers and mild, wet winters. The landscape is dominated by irrigated permanent crops (orchards, vineyards) and annual field crops. The simulation period (April--June) captures spring green-up and early crop development, while the ground truth period (July--October) spans peak summer productivity through early senescence. The strong phenological signal and relatively cloud-free conditions make this site suitable for the standard  SENDAI Jr. architecture.

\subsection{Midwestern United States: Iowa Corn Belt}

The Iowa Corn Belt exemplifies temperate continental agriculture (K\"oppen Dfa) with pronounced seasonal temperature variation. Land cover is dominated by corn--soybean rotations under rainfed conditions. The April--June simulation period captures emergence and vegetative growth, while July--October encompasses reproductive stages through harvest. Despite the dramatic phenological transition, the relatively predictable cropping calendar and absence of irrigation artifacts allow reconstruction with standard SENDAI Jr.

\subsection{Southwestern United States: Imperial Valley, California}

Imperial Valley presents an extreme hot desert environment (K\"oppen BWh) where agriculture is entirely dependent on irrigation from the Colorado River. The sharp contrast between verdant irrigated fields and surrounding bare desert creates pronounced spatial heterogeneity. Multiple cropping cycles per year and variable irrigation schedules introduce high-frequency temporal dynamics requiring the full SENDAI architecture to capture sub-seasonal variability.

\subsection{Western Spain: Lower Guadalquivir Valley}

The Guadalquivir Valley exhibits Mediterranean climate (K\"oppen Csa) with a distinctive reversed phenological calendar compared to northern hemisphere temperate regions---vegetation green-up occurs during mild, wet winters rather than spring. The February--April simulation period captures peak winter greenness and early spring drying, while September--December represents the onset of the growing season following summer drought. This phenological inversion provides a critical test of domain adaptation capability.



\subsection{Northwestern China: Tarim Basin}

The Tarim Basin represents a hyper-arid cold desert environment (K\"oppen BWk) surrounded by the Tianshan and Kunlun mountain ranges. Agriculture is concentrated in narrow oasis strips fed by glacial meltwater, creating extreme spatial gradients between irrigated fields and surrounding desert. The April--June simulation period captures spring irrigation onset and crop establishment, while July--October encompasses peak productivity. The localized nature of vegetation and strong background contrasts require hierarchical frequency decomposition.

\subsection{Southeastern Australia: Riverina Region}

The Riverina represents Australia's ``food bowl'', featuring humid subtropical climate (K\"oppen Cfa) transitional to semi-arid conditions. Mixed cropping systems include irrigated rice, wheat, and pasture. The Southern Hemisphere location provides reversed seasonality, with February--April capturing late summer/autumn conditions and September--December spanning spring green-up. This site tests generalization to different hemispheric phenological timing and drought-prone conditions.



\section{Seasonal Split Strategy and Sparse Sensor Configuration Details}
\label{app:sensor_config}

Table~\ref{tab:study_sites} presents study site characteristics and observation periods, motivated by phenological domain shift. Together, they exhibit phenological coherence but heterogeneous spatiotemporal transitions in NDVI: vegetation dynamics vary not only across space but also exhibit distinct temporal autocorrelation structures between seasons (due to land cover heterogeneity, soil moisture gradients, and management practices, etc). For Southern Hemisphere sites and Mediterranean climates with reversed seasonality, the seasonal splits capture analogous phenological transitions adapted to local climate rhythms. The SENDAI framework must therefore learn to reconstruct fields that are simultaneously spatially heterogeneous and temporally non-stationary.

A central tenet of the SENDAI architecture is that full-state reconstruction can be achieved from sparse temporal observations at a limited number of spatial locations, leveraging Takens' embedding theorem and learned decoder representations~\citep{williams2024sensing, bao2025data}. We employ $p = 64$ randomly placed sensors, representing 1.56\% of the full state space ($64 \times 64 = 4096$ pixels)---comparable to or lower than typical ground-based monitoring network densities. Sensor locations are randomly selected, excluding boundary pixels (2-pixel buffer). Following the SHRED formulation, sensor observations are organized into time-delay embeddings of length $L = 5$ lags, where at each time step $t$, the input to the encoder consists of the sensor history $\mathbf{S}_t = [\mathbf{s}_{t-L+1}, \ldots, \mathbf{s}_t]^\top \in \mathbb{R}^{L \times p}$, enabling the model to infer temporal derivatives and phenological trends critical for tracking dynamic vegetation changes.

\section{Baseline Implementation Details}
\label{app:baselines}

\subsection{SG + IDW Implementation}

Savitzky-Golay filtering is applied with window length 7 and polynomial order 2, providing local temporal smoothing while preserving phenological trends. Inverse Distance Weighting uses power parameter $p = 2$ (standard inverse-square weighting). This baseline represents the computational floor---no training required, sub-second inference.

\subsection{HANTS + IDW Implementation}

HANTS fits a truncated Fourier series with 3 harmonic terms (annual, semi-annual, and quarterly cycles) at each sensor location. The fitting procedure uses iterative refinement to reject cloud-contaminated observations. Spatial interpolation follows the same IDW protocol as SG+IDW.

\subsection{Kriging Implementation}

Gaussian Process regression employs an RBF (squared exponential) kernel with automatic relevance determination. For Kriging-GT, lengthscale and variance hyperparameters are optimized via maximum likelihood on the ground truth sensor observations at each timestep. For Kriging-Sim, hyperparameters are pre-fitted on simulation full fields and held fixed during ground truth reconstruction, providing an intentionally favorable setting analogous to the use of simulation for encoder-decoder pretraining.

\subsection{MMGN Implementation}
\label{app:mmgn}

The Multiplicative and Modulated Gabor Network (MMGN)~\citep{luo2024continuous} represents a recent advance in implicit neural representations (INRs) for continuous field reconstruction from sparse observations. We implement MMGN following the original architecture and training protocol to provide a rigorous comparison with neural network-based approaches.

\paragraph{Architecture.} MMGN employs an auto-decoder architecture that learns a continuous spatial representation conditioned on per-timestep latent codes. The decoder consists of multiplicative layers combining Gabor filters with bilinear fusion of coordinates and latent information:
\begin{equation}
\mathbf{h}^{(0)} = g_0(\mathbf{x}) \odot \mathcal{F}_0(\mathbf{0}, \mathbf{z}), \quad
\mathbf{h}^{(\ell+1)} = g_{\ell+1}(\mathbf{x}) \odot \mathcal{F}_{\ell+1}(\mathbf{h}^{(\ell)}, \mathbf{z}),
\end{equation}
where $g_\ell(\mathbf{x})$ are Gabor filters applied to spatial coordinates, $\mathcal{F}_\ell$ are bilinear fusion layers, and $\mathbf{z} \in \mathbb{R}^{d_z}$ is a learnable latent code for each time instance. The Gabor filters take the form:
\begin{equation}
g_\ell(\mathbf{x}) = \sin(\mathbf{W}_g \mathbf{x} + \mathbf{b}_g) \odot \exp\left(-\frac{\gamma}{2}\|\mathbf{x} - \boldsymbol{\mu}\|^2\right),
\end{equation}
with learnable centers $\boldsymbol{\mu}$ and bandwidth parameters $\gamma$ sampled from a Gamma distribution. This multiplicative structure enables the network to represent the output as a linear combination of Gabor basis functions, providing shift-invariance properties beneficial for spatial field reconstruction.

\paragraph{Adaptation to Our Setting.} The original MMGN was evaluated on climate simulation data (CESM2 global surface temperature, $192 \times 288$ resolution, 1024 timesteps) and satellite sea surface temperature (GHRSST, $901 \times 1001$ resolution, 360 timesteps) with sensor coverage ranging from 5\% to 50\%. Our MODIS NDVI reconstruction task presents four key differences: (i) substantially sparser observations (1.56\% vs.\ 5--50\%), (ii) smaller spatial extent ($64 \times 64$ pixels), (iii) significantly shorter temporal sequences ($\sim$70 timesteps vs.\ 360--1024), and (iv) heterogeneous agricultural landscapes with sharp boundaries rather than smoothly varying oceanographic or atmospheric fields.

We apply MMGN with identical sensor configurations (64 randomly placed sensors) and temporal splits as all other methods. The model learns separate latent codes for each timestep, with training performed on sensor observations only and evaluation on full-field reconstruction.



\section{SENDAI Architecture Details}
\label{app:architecture}

This appendix provides complete architectural specifications for the SENDAI components discussed in Section~\ref{sec:method}.

\subsection{Base SHRED Model}
\label{app:shred}

The base SHRED architecture consists of an LSTM temporal encoder and an MLP spatial decoder.

\paragraph{LSTM Encoder.} The encoder processes sensor time-histories $\mathbf{S}_{t-L:t} \in \mathbb{R}^{L \times p}$:
\begin{equation}
\mathbf{h}_\tau, \mathbf{c}_\tau = \text{LSTM}(\mathbf{s}_\tau, \mathbf{h}_{\tau-1}, \mathbf{c}_{\tau-1}), \quad \tau = t-L+1, \ldots, t,
\label{eq:lstm_update}
\end{equation}
with $K$ stacked LSTM layers (typically $K=2$), hidden dimension $d_z$, and dropout applied between layers during training. The latent representation is:
\begin{equation}
\mathbf{z} = \text{LayerNorm}(\mathbf{h}_t^{(K)}),
\end{equation}
where $\mathbf{h}_t^{(K)}$ is the final hidden state of the top layer.

\paragraph{MLP Decoder.} The decoder maps latent codes to full spatial states:
\begin{equation}
\mathcal{D}_{\text{LF}}(\mathbf{z}) = \mathbf{W}_D \cdot \text{ReLU}(\text{LN}(\mathbf{W}_{D-1} \cdots \text{ReLU}(\text{LN}(\mathbf{W}_1 \mathbf{z} + \mathbf{b}_1)) \cdots)),
\end{equation}
where LN denotes layer normalization and the hidden layer dimensions are specified in Table~\ref{tab:hyperparams}.

\subsection{DA-SHRED Latent Transform}
\label{app:dashred}

The latent transformation module adapts simulation-trained representations to ground truth sensor data:
\begin{equation}
\mathcal{T}(\mathbf{z}) = \tanh\left(\mathbf{W}_2 \cdot \text{ReLU}(\mathbf{W}_1 \mathbf{z} + \mathbf{b}_1) + \mathbf{b}_2\right),
\end{equation}
with the transformed latent given by Eq.~\eqref{eq:gan_alignment}. The $\tanh$ nonlinearity bounds the correction magnitude, and the learnable scale $\gamma$ is initialized to 0.1.

\paragraph{GAN Discriminator.} The discriminator $\mathcal{D}: \mathbb{R}^{d_z} \to [0,1]$ is a 3-layer MLP with LeakyReLU activations (slope 0.2):
\begin{equation}
\mathcal{D}(\mathbf{z}) = \sigma\left(\mathbf{W}_3 \cdot \text{LReLU}(\mathbf{W}_2 \cdot \text{LReLU}(\mathbf{W}_1 \mathbf{z}))\right),
\end{equation}
where $\sigma$ is the sigmoid function. Training uses the binary cross-entropy objectives:
\begin{align}
\mathcal{L}_D &= -\mathbb{E}_{\mathbf{z} \sim p_{\text{gt}}}[\log \mathcal{D}(\mathbf{z})] - \mathbb{E}_{\mathbf{z} \sim p_{\text{sim}}}[\log(1 - \mathcal{D}(\mathcal{G}(\mathbf{z})))], \\
\mathcal{L}_G &= -\mathbb{E}_{\mathbf{z} \sim p_{\text{sim}}}[\log \mathcal{D}(\mathcal{G}(\mathbf{z}))].
\end{align}

\paragraph{SENDAI Jr. Pipeline.}
\label{app:simplified_details}
For datasets where the sim2real gap is primarily low-frequency, a simplified two-stage pipeline suffices. Stage 1 trains the LSTM encoder and MLP decoder on simulation data with either sparse sensor histories or full-state supervision until convergence. Stage 2 then encodes both simulation and ground truth sensor data to obtain latent distributions and trains the GAN components (generator $\mathcal{G}$ and discriminator $\mathcal{D}$) on these latent codes to align the distributions. This variant relies solely on latent-space alignment to bridge the sim2real gap and is recommended as a baseline before deploying the full hierarchical architecture. The full pipeline should be used when spectral analysis of post-alignment residuals reveals significant high-frequency structure, evidenced by distinct peaks in the FFT magnitude spectrum of $\mathbf{s}'(t) - \mathbf{M}\tilde{\mathbf{u}}_{\text{LF}}(t)$.

\subsection{Coordinate-Based INR for HF Correction}
\label{app:inr_details}

The implicit neural representation for high-frequency correction consists of an encoder and coordinate-based decoder.

\paragraph{Sensor Residual Encoder.} Maps $p$-dimensional residuals to latent codes:
\begin{equation}
\mathcal{E}_{\text{HF}}(\mathbf{r}) = \mathbf{W}_E^{(2)} \cdot \text{ReLU}(\text{LN}(\mathbf{W}_E^{(1)} \mathbf{r} + \mathbf{b}_E^{(1)})) + \mathbf{b}_E^{(2)},
\end{equation}
with output dimension $d_{\text{HF}}$ (typically 64).

\paragraph{Fourier Positional Encoding.} For coordinates $(x, y) \in [0,1]^2$ and $L$ frequency bands with log-spaced frequencies $\sigma_j = 2^{(j-1) \log_2 \sigma_{\max} / (L-1)}$:
\begin{equation}
\text{PE}(x, y) = \left[x, y, \sin(2\pi\sigma_1 x), \cos(2\pi\sigma_1 x), \sin(2\pi\sigma_1 y), \cos(2\pi\sigma_1 y), \ldots\right] \in \mathbb{R}^{2 + 4L}.
\end{equation}
Typical values are $L = 16$ frequency bands with $\sigma_{\max} = 8.0$.

\paragraph{Coordinate Decoder.} The decoder MLP takes the concatenation $[\text{PE}(x,y); \mathbf{z}_{\text{HF}}] \in \mathbb{R}^{2+4L+d_{\text{HF}}}$:
\begin{equation}
\mathcal{D}_{\text{INR}}([\text{PE}; \mathbf{z}]) = \mathbf{W}_3 \cdot \text{ReLU}(\text{LN}(\mathbf{W}_2 \cdot \text{ReLU}(\text{LN}(\mathbf{W}_1 [\text{PE}; \mathbf{z}])))).
\end{equation}
The final layer produces a scalar output for each queried coordinate.

\paragraph{Batched Coordinate Queries.} At inference, all $n = H \times W$ grid coordinates are queried simultaneously. For memory efficiency with large grids, coordinates are processed in chunks:
\begin{equation}
u_{\text{HF}}(i, j) = \gamma \cdot \mathcal{D}_{\text{INR}}\left(\left[\text{PE}(i/H, j/W); \mathbf{z}_{\text{HF}}\right]\right), \quad \forall (i,j) \in \{0,\ldots,H-1\} \times \{0,\ldots,W-1\}.
\end{equation}

\subsection{2D Frequency Sparsity Regularization}
\label{app:sparsity_details}

For 2D spatial fields, frequency sparsity is computed via the 2D real FFT.

\paragraph{Frequency Grid.} Let $\hat{\mathbf{u}} = \text{rfft2}(\mathbf{u})$ with shape $(H, W/2+1)$. The frequency coordinates are:
\begin{align}
k_y &\in \{0, 1, \ldots, H/2, -H/2+1, \ldots, -1\}, \\
k_x &\in \{0, 1, \ldots, W/2\}.
\end{align}
The frequency radius is $\|\mathbf{k}\| = \sqrt{k_y^2 + k_x^2}$.

\paragraph{Bandlimited Sparsity.} The in-band region $\mathcal{B} = \{\mathbf{k}: \|\mathbf{k}\| \leq k_{\max}\}$ and out-of-band $\bar{\mathcal{B}} = \{\mathbf{k}: \|\mathbf{k}\| > k_{\max}\}$:
\begin{equation}
\mathcal{R}_{\text{band}}(\mathbf{u}) = \frac{\sum_{\mathbf{k} \in \mathcal{B}} |\hat{u}_\mathbf{k}|}{\sqrt{\sum_{\mathbf{k} \in \mathcal{B}} |\hat{u}_\mathbf{k}|^2 + \epsilon}} + \beta_1 \cdot \frac{\sum_{\mathbf{k} \in \bar{\mathcal{B}}} |\hat{u}_\mathbf{k}|^2}{\sum_{\mathbf{k}} |\hat{u}_\mathbf{k}|^2 + \epsilon}.
\end{equation}

\paragraph{Frequency Exclusion.} For layer $\ell$, let $\{(\bar{k}_y^{(j)}, \bar{k}_x^{(j)})\}_{j=1}^{J}$ be frequencies discovered by previous layers. The exclusion region with radius $r_{\text{exc}}$ is:
\begin{equation}
\mathcal{E}^{(\ell)} = \bigcup_{j=1}^{J} \left\{\mathbf{k}: \sqrt{(k_y - \bar{k}_y^{(j)})^2 + (k_x - \bar{k}_x^{(j)})^2} < r_{\text{exc}}\right\}.
\end{equation}
Due to conjugate symmetry of real signals, both $(k_y, k_x)$ and $(-k_y, k_x)$ are excluded. The exclusion penalty uses weight $\beta_2$:
\begin{equation}
\mathcal{P}_{\mathcal{E}}^{(\ell)} = \beta_2 \cdot \frac{\sum_{\mathbf{k} \in \mathcal{E}^{(\ell)}} |\hat{u}_\mathbf{k}|^2}{\sum_{\mathbf{k}} |\hat{u}_\mathbf{k}|^2 + \epsilon}.
\end{equation}
After training each HF peeling layer, dominant frequencies are identified by computing the 2D FFT of the mean HF output over the training set, excluding the DC component, and extracting the top-$k_\ell$ unique frequency locations accounting for conjugate symmetry. These are then added to the exclusion set for subsequent layers.

\paragraph{Top-$k_\ell$ Sparsity.} For peeling layer $\ell$ with adaptively selected $k_\ell$ modes:
\begin{equation}
\mathcal{R}_{\text{topk}}(\mathbf{u}; k_\ell) = 1 - \frac{\sum_{i=1}^{k_\ell} |\hat{u}_{(i)}|^2}{\sum_{\mathbf{k}} |\hat{u}_\mathbf{k}|^2 + \epsilon},
\end{equation}
where $|\hat{u}_{(i)}|$ is the $i$-th largest Fourier magnitude.

\begin{table}[t]
\centering
\caption{Default hyperparameters for the SENDAI architecture.}
\label{tab:hyperparams}
\begin{tabular}{llc}
\toprule
\textbf{Component} & \textbf{Parameter} & \textbf{Value} \\
\midrule
LSTM Encoder & Hidden dimension $d_z$ & 32 \\
 & Number of layers $K$ & 2 \\
 & Dropout rate & 0.1 \\
 & Temporal lags $L$ & 5 \\
\midrule
LF Decoder & Hidden layers & [256, 256] \\
 & Activation & ReLU \\
\midrule
GAN & Generator hidden & 64 \\
 & Discriminator hidden & 64 \\
\midrule
INR (per HF layer) & Latent dimension $d_{\text{HF}}$ & 64 \\
 & Encoder hidden & [128, 128] \\
 & Decoder hidden & [256, 256, 128] \\
 & PE frequencies $L$ & 16 \\
 & PE max frequency $\sigma_{\max}$ & 8.0 \\
 & Scale $\gamma$ init & 0.1 \\
\midrule
HF Training & Warmup epochs $E_{\text{warm}}$ & 100 \\
 & $\lambda_{\text{sp}}$ & 0.05 \\
 & $\lambda_{\text{sm}}$ & 0.1 \\
 & Fine-tune $\lambda'_{\text{sp}}$ & 0.005 \\
\midrule
Sparsity Penalties & Out-of-band $\beta_1$ & 100 \\
 & Exclusion $\beta_2$ & 100 \\
 & Top-k weight $\lambda_{\text{topk}}$ & 10.0 \\
 & Exclusion radius $r_{\text{exc}}$ & 2.0 \\
\midrule
Adaptive $k_\ell$ & Bandwidth tolerance $\Delta k$ & 2.0 \\
 & Energy threshold $\rho$ & 0.8 \\
\midrule
Optimization & Learning rate & $10^{-4}$ \\
 & Batch size & 16 \\
 & Optimizer & AdamW \\
\bottomrule
\end{tabular}
\end{table}

\subsection{Smoothness Regularization Options}
\label{app:smoothness}

Three smoothness regularization strategies are supported, with selection guided by prior knowledge of the expected HF field characteristics:

\paragraph{Gradient (Total Variation).} For fields with sharp edges or discontinuities:
\begin{equation}
\mathcal{R}_{\text{grad}}(\mathbf{u}) = \frac{1}{HW}\sum_{i,j} \left[(u_{i,j+1} - u_{i,j})^2 + (u_{i+1,j} - u_{i,j})^2\right].
\end{equation}
This penalizes gradients uniformly, promoting piecewise constant solutions.

\paragraph{Laplacian (Curvature).} For fields expected to be smooth with gentle variations:
\begin{equation}
\mathcal{R}_{\text{lap}}(\mathbf{u}) = \frac{1}{(H-2)(W-2)}\sum_{i,j} \left[u_{i+1,j} + u_{i-1,j} + u_{i,j+1} + u_{i,j-1} - 4u_{i,j}\right]^2.
\end{equation}
This penalizes curvature (second derivatives) rather than gradients, allowing linear ramps and sharp but smooth features while suppressing high-frequency oscillations.

\paragraph{Bilateral (Edge-Preserving).} For fields with both smooth regions and sharp boundaries:
\begin{equation}
\mathcal{R}_{\text{bilateral}}(\mathbf{u}) = \sum_{i,j} \left[\text{Huber}_\delta(u_{i,j+1} - u_{i,j}) + \text{Huber}_\delta(u_{i+1,j} - u_{i,j})\right],
\end{equation}
where $\text{Huber}_\delta(x) = \frac{1}{2}x^2$ if $|x| < \delta$, else $\delta(|x| - \frac{\delta}{2})$. The threshold $\delta$ controls the transition between quadratic (small gradients) and linear (large gradients) penalization, preserving edges while smoothing homogeneous regions.

\subsection{Hyperparameter Configuration}
\label{app:hyperparams}

Table~\ref{tab:hyperparams} summarizes the default hyperparameters.

\begin{table}[t]
\centering
\caption{Computational cost and model complexity comparison.}
\label{tab:computation}
\begin{tabular}{lc|lc}
\toprule
\multicolumn{2}{c|}{\textbf{SENDAI Jr.}} & \multicolumn{2}{c}{\textbf{SENDAI}} \\
\midrule
\multicolumn{2}{l|}{\textit{Model Complexity}} & \multicolumn{2}{l}{\textit{Model Complexity}} \\
\quad SHRED (LSTM + Decoder) & 1,149.0 K & \quad SHRED (LSTM + Decoder) & 1,149.0 K \\
\quad DA Transform & 4.2 K & \quad DA Transform & 4.2 K \\
 & & \quad HF Peeling Layers & 334.5 K \\
\quad \textbf{Total parameters} & \textbf{1,153.2 K} & \quad \textbf{Total parameters} & \textbf{1,487.7 K} \\
\midrule
\multicolumn{2}{l|}{\textit{Training Time}} & \multicolumn{2}{l}{\textit{Training Time}} \\
\quad Stage 1 (SHRED) & 2.76 sec & \quad Stage 1 (SHRED) & 2.85 sec \\
\quad Stage 2 (DA-SHRED + GAN) & 10.26 sec & \quad Stage 2 (DA-SHRED + GAN) & 14.18 sec \\
 & & \quad Stage 3 (Hierarchical HFP + INR) & 25 min 50 sec \\
\quad \textbf{Total} & \textbf{15.94 sec} & \quad \textbf{Total} & \textbf{26 min 12 sec} \\
\midrule
\multicolumn{4}{c}
{\textit{Hardware}} \\
\multicolumn{4}{c}{\quad CPU: Apple M4, Memory: 24 GB} \\
\bottomrule
\end{tabular}
\end{table}

\subsection{Computational Cost and Model Complexity}
\label{app:complexity}

Table~\ref{tab:computation} summarizes the computational cost and model complexity.

\section{Synthetic Validation: Extended Details}
\label{app:synthetic}

This appendix provides comprehensive details for the synthetic validation experiments presented in Section~\ref{sec:synthetic}.

\subsection{Traveling Wave System: Data Generation}
\label{app:synthetic_system}


The synthetic traveling wave system is generated on a spatial domain $x \in [0, 2\pi]$ with $N=128$ grid points and temporal domain $t \in [0, 10]$ with $\Delta t = 0.05$ (200 timesteps). The full three-mode field is:
\begin{equation}
u(x,t) = \sin(2x - t) + 0.4\sin(5x - 3t) + 0.25\sin(11x - 7t),
\end{equation}
where the different temporal frequencies $\omega_1=1$, $\omega_2=3$, $\omega_3=7$ ensure the three modes remain distinguishable as the system evolves. The simulation model contains only the first term, representing a simplified physics model that misses intermediate and fine-scale dynamics.

\subsection{NDVI Slice Extraction}
\label{app:ndvi_slice_setup}

For the NDVI slice experiment, we extract a 1D transect from the Tarim Basin site at $x = 25\%$ of the image width (column 16 of a $64 \times 64$ grid). This slice traverses a sharp mountain-basin boundary, capturing the heterogeneous vegetation structure characteristic of the landscape.

Both simulation data from the April--June period and ground truth from July--October are aligned to $T=72$ timesteps for training. The HF residual exhibits:
\begin{itemize}
    \item Range: $[-0.46, 0.86]$ NDVI units
    \item Standard deviation: 0.212
    \item Sharp discontinuities at mountain boundaries (pixels 10--20, 45--55)
    \item Temporal variability from phenology
\end{itemize}

\subsection{Sensor Configuration}
\label{app:synthetic_sensors}

Both experiments employ an extremely sparse configuration of only $p=3$ sensors. For the traveling wave system, this provides approximately 2.3\% spatial coverage of the 128-point grid. For the NDVI slice, sensors are randomly placed as well. Despite this severe undersampling, both experiments successfully recover the missing frequency content, demonstrating the framework's ability to exploit temporal coherence for frequency recovery.

The LSTM encoder processes temporal histories of length $L=20$ lags (traveling wave) or $L=10$ lags (NDVI slice), exploiting temporal coherence to compensate for spatial sparsity.

For the NDVI experiment, INR decoders with Fourier positional encoding are used to produce spatially coherent outputs despite the sharp discontinuities in the target field.

\subsection{Joint vs. Hierarchical Frequency Discovery}
\label{app:synthetic_comparison}

\begin{figure}[t]
    \centering
    \includegraphics[width=\columnwidth]{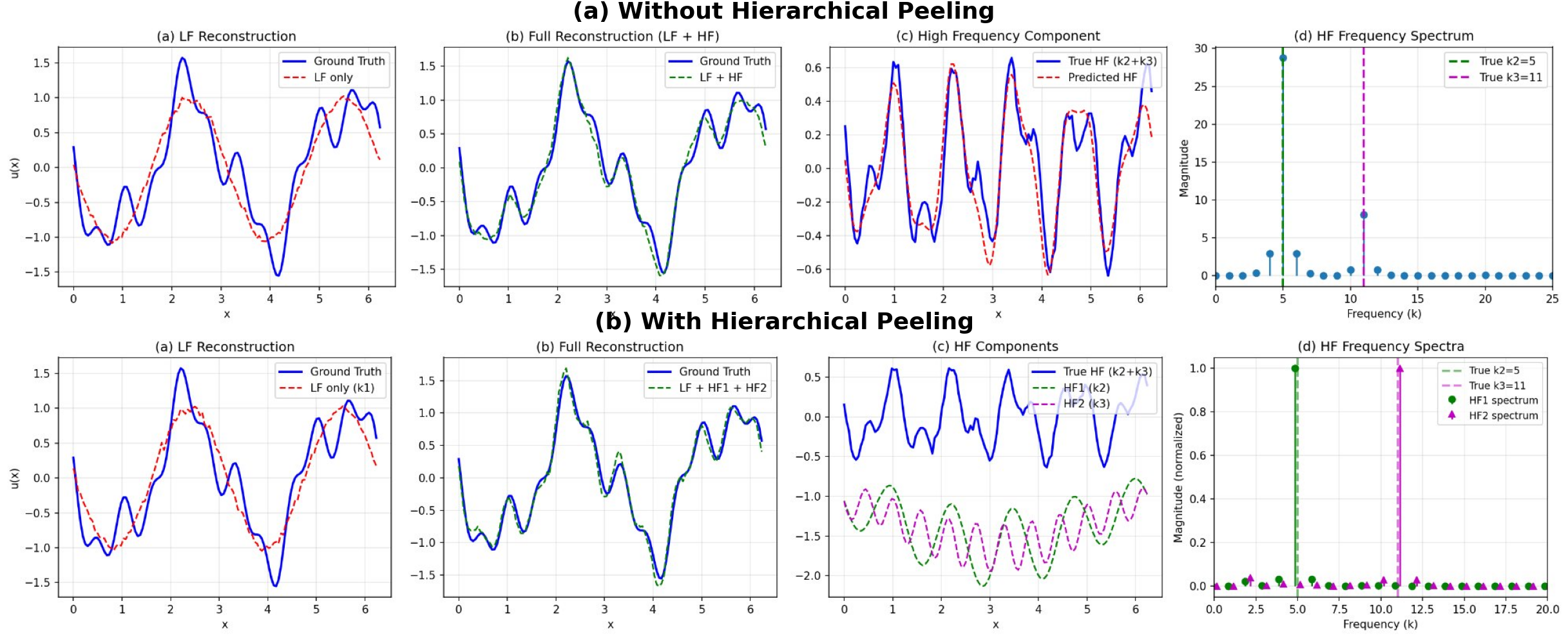}
    \caption{Comparison of joint and hierarchical frequency discovery on the three-mode traveling wave system for a single time-point reconstruction. \textbf{(a)} the frequency spectrum shows energy leakage to non-target modes. \textbf{(b)} modes are discovered sequentially, yielding spectrally clean outputs and improved fine-scale fidelity. In third panel, HF1 and HF2 show the individually learned components (unnormalized); their sum after scaling recovers the true HF.}
    \label{fig:synthetic_comparison}
\end{figure}

\begin{figure}[t]
    \centering
    \includegraphics[width=0.5\textwidth]{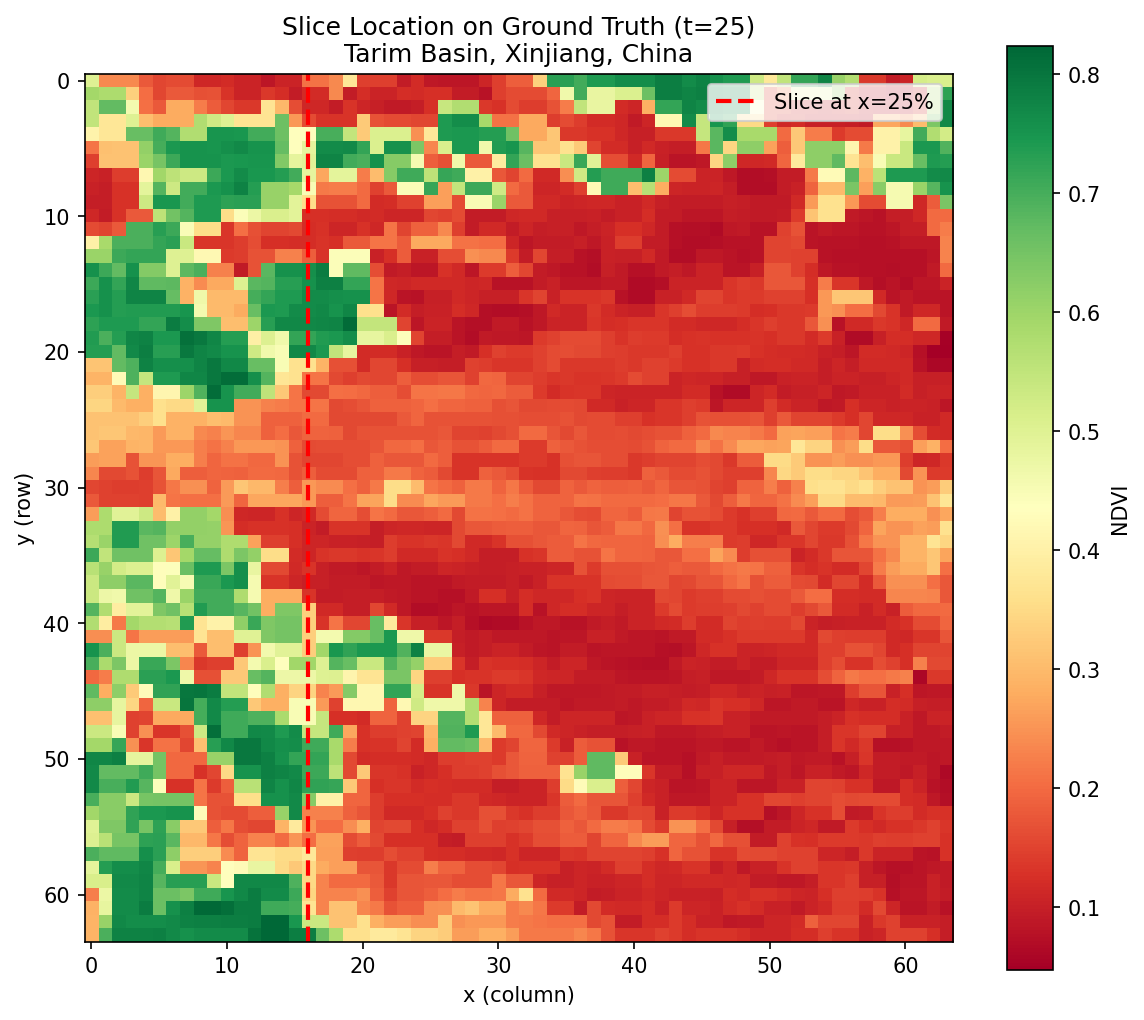}
    \caption{Slice location for hierarchical frequency peeling on a 1D NDVI slice from the Tarim Basin.}
    \label{fig:ndvi_slice}
\end{figure}

We compare two strategies for HF correction on the traveling wave system:

\paragraph{Joint Discovery.} A single HF pathway with bandlimited sparsity discovers both $k_2$ and $k_3$ simultaneously. While achieving 84.7\% RMSE improvement, the frequency spectrum shows energy spread across modes $k \in \{3, 4, 5, 10, 11, 12\}$ rather than concentrated solely at the targets. The combined HF output represents an entangled mixture with energy leakage.

\paragraph{Sequential Peeling.} The hierarchical approach achieves 85.1\% RMSE improvement with clean spectrum: HF$_1$ captures $k=5$ with $>95\%$ of its energy at the target mode, and HF$_2$ captures $k=11$ similarly. This spectral purity enables physical interpretation of individual frequency contributions, downstream analysis of specific spectral components, and modular addition of peeling layers without retraining.

\subsection{NDVI Transect Decomposition}
The 1D NDVI slice from the Tarim Basin site illustrates how hierarchical frequency peeling decomposes a complex, noisy simulation--observation residual into interpretable components across scales.

\paragraph{HF$_1$: Climate-Driven Seasonal Dynamics.}
The first peeling layer captures the dominant residual variability, combining coherent temporal fluctuations with a broad spatial gradient along the transect. Its synchronized, episode-like oscillations are consistent with basin-wide meteorological forcing (e.g., temperature anomalies and intermittent precipitation events) acting across the slice, while the gradual late-season attenuation suggests a phenological convergence between spring-calibrated simulations and summer--autumn observations. The spatial amplitude structure further indicates modulation by the elevation gradient, supporting an interpretation of temperature-mediated phenological offsets that vary systematically across landscape zones.

\paragraph{HF$_2$: Hydrological Persistence.}
The second peeling layer is characterized by stronger, more persistent spatial stratification with relatively minimal temporal evolution, indicating a quasi-stationary control on the residual. In the hyperarid Tarim Basin, such stable structure could be plausibly explained by differential water availability governed by proximity to snow-melt, groundwater access, and topographic accumulation of runoff. 

\paragraph{HF$_3$: Edaphic and Microsite Heterogeneity.}
The third peeling layer isolates finer-scale, temporally invariant spatial heterogeneity, pointing to localized controls that persist over the season. We attribute this component to edaphic and microsite factors such as soil texture, salinity, nutrient level, and geomorphic microhabitats (e.g., abandoned channels or alluvial features) that influence vegetation independently of broad climate and hydrological gradients. Predominantly negative anomalies are consistent with systematic overestimation by the low-frequency reconstruction in sub-pixel heterogeneous patches that cannot be resolved without an explicit multi-scale residual model.

\subsection{Advantages of Hierarchical Peeling}
\label{app:synthetic_advantages}

The experiments reveal several advantages of hierarchical peeling:

\paragraph{Robustness to Non-Ideal Signals.} The NDVI slice experiment demonstrates that hierarchical peeling succeeds even when the target signal doesn't have clean sinusoidal patterns. The landscape creates sharp discontinuities (effectively high bandwidth), yet the peeling layers correctly identify the majority of spectral modes without being corrupted by edge effects.

\paragraph{Frequency Exclusion Mechanism.} The exclusion penalty prevents mode leakage where a subsequent layer partially recaptures previously discovered content. In the NDVI experiment, HF$_2$ correctly discovers $k=4$ rather than reinforcing the $k=2$ mode already captured by HF$_1$, despite $k=2$ having substantial residual energy.

\paragraph{Stable Training Dynamics.} By constraining each layer to discover a concentration of modes, the training procedure becomes more robust. The traveling wave experiment shows HF$_1$ converging to $k=5$ before epoch 200, while HF$_2$ locks onto $k=11$ shortly after. The NDVI experiment exhibits similar sequential convergence despite the noisy target signal.

\paragraph{Interpretability.}
The adaptive HF selection encourages each layer to focus on the most salient remaining structure, while the exclusion mechanism reduces redundancy across layers, yielding a nested decomposition that maps naturally onto scale-dependent control mechanisms.





\subsection{Implications for Full 2D Reconstruction}
\label{app:synthetic_implications}

The synthetic validation confirms two key capabilities essential for the full SENDAI framework:

\paragraph{Extreme Sparsity Tolerance.} Both experiments succeed with only $p=3$ sensors ($\sim$2--5\% coverage), demonstrating the framework's ability to exploit temporal coherence for frequency recovery. This directly supports the NDVI reconstruction task where 64 sensors cover only 1.56\% of the spatial domain.

\paragraph{Universality Across Signal Types.} The success on both clean synthetic waves and noisy real NDVI data---with outliers, sharp discontinuities, and non-stationary dynamics---demonstrates that hierarchical peeling is not limited to idealized signals. The frequency exclusion mechanism and INR decoders together handle the heterogeneous spatiotemporal fields characteristic of real remote sensing applications.

These findings motivate the hierarchical architecture employed in the full SENDAI framework, where multiple peeling layers with coordinate-based INR decoders sequentially extract interpretable spectral corrections from the sim2real discrepancy.

\section{SENDAI Jr. Site-Specific Results}
\label{app:dashred_details}

This appendix provides detailed analyses and qualitative reconstruction results for the three sites evaluated using the SENDAI Jr. pipeline. Performance is assessed using both RMSE and the Structural Similarity Index Measure (SSIM), with SSIM serving as the primary indicator of reconstruction quality. As explained in the main text, baseline methods fail to preserve the topological structure of spatial patterns despite achieving moderate RMSE values. IDW-based methods exhibit ``bullseye'' artifacts centered at sensor locations. Kriging produces overly smooth reconstructions that obscure sharp boundaries. SSIM captures the preservation of spatial patterns, textures, and structural information that RMSE alone cannot assess---critically important for remote sensing applications where topological fidelity determines downstream analysis utility.

\subsection{Central Valley, California, USA}

This irrigated cropland site achieves RMSE of 0.1068 and SSIM of 0.5747, representing a 120\% SSIM improvement over the best-performing baseline (SG+IDW at 0.2612). This substantial structural improvement indicates that SENDAI Jr. successfully preserves field boundaries, irrigation patterns, and vegetation gradients that baseline methods systematically destroy.

\begin{figure}[t]
\centering
\includegraphics[width=\columnwidth]{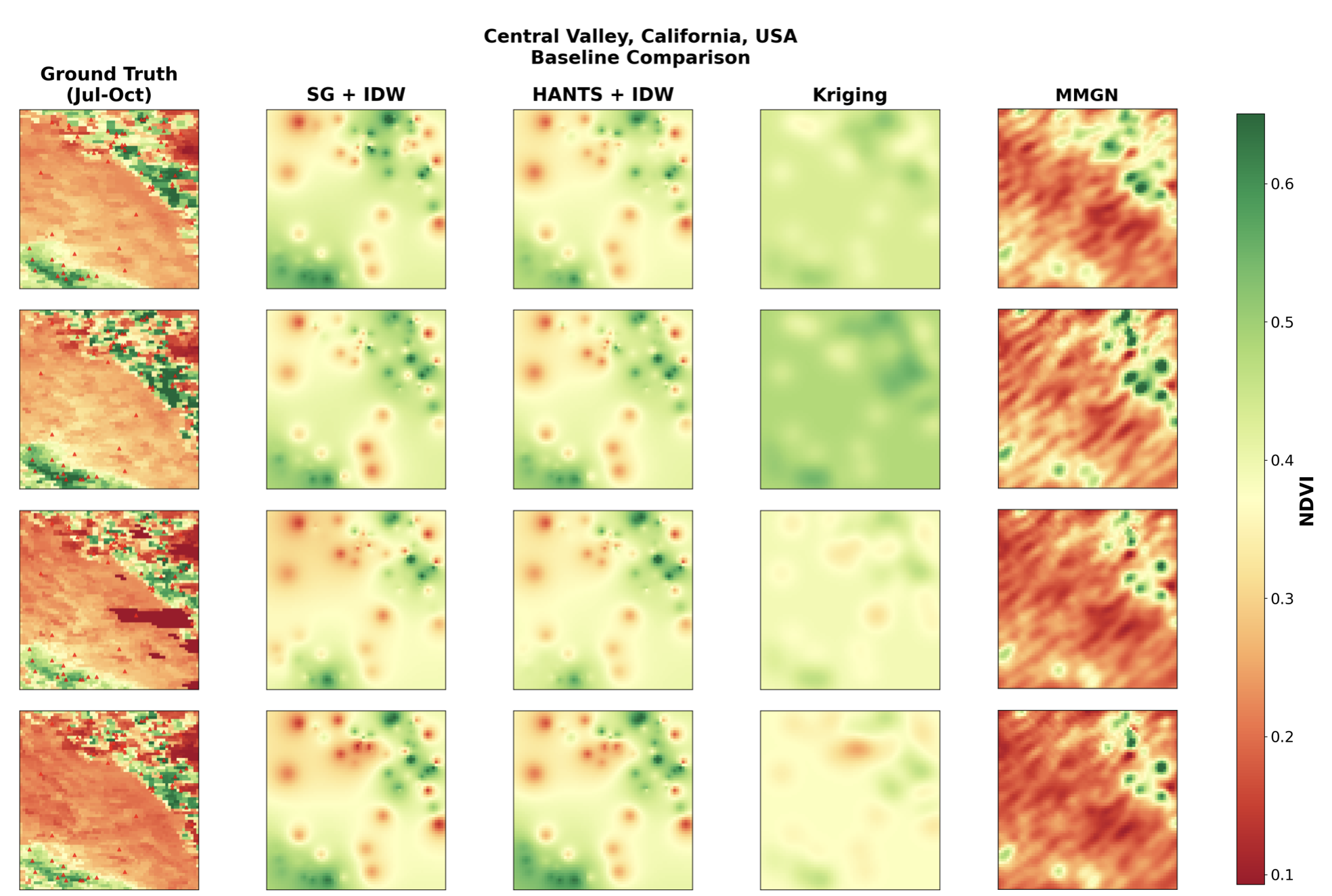}
\caption{Baseline reconstruction comparison for Central Valley, California. Each row represents an equally-spaced temporal frame within the ground truth period. The irrigated cropland landscape exhibits field-scale heterogeneity that baseline interpolation methods fail to faithfully reproduce. Note the bullseye artifacts in IDW-based methods and over-smoothing in Kriging, which are reflected in their low SSIM values.}
\label{fig:baseline_cv_app}
\end{figure}

Figure~\ref{fig:baseline_cv_app} shows baseline performance on the Central Valley site. While this site exhibits less extreme heterogeneity than the Tarim Basin, the baseline methods still produce characteristic artifacts: IDW-based methods show bullseye patterns around sensor locations (SSIM: 0.2612 and 0.2504), while Kriging over-smooths field boundaries resulting in the lowest SSIM (0.0922). The poor SSIM values of baseline methods---despite moderate RMSE---demonstrate the critical importance of structural similarity metrics for evaluating reconstruction quality.

\begin{figure}[t]
\centering
\includegraphics[width=\columnwidth]{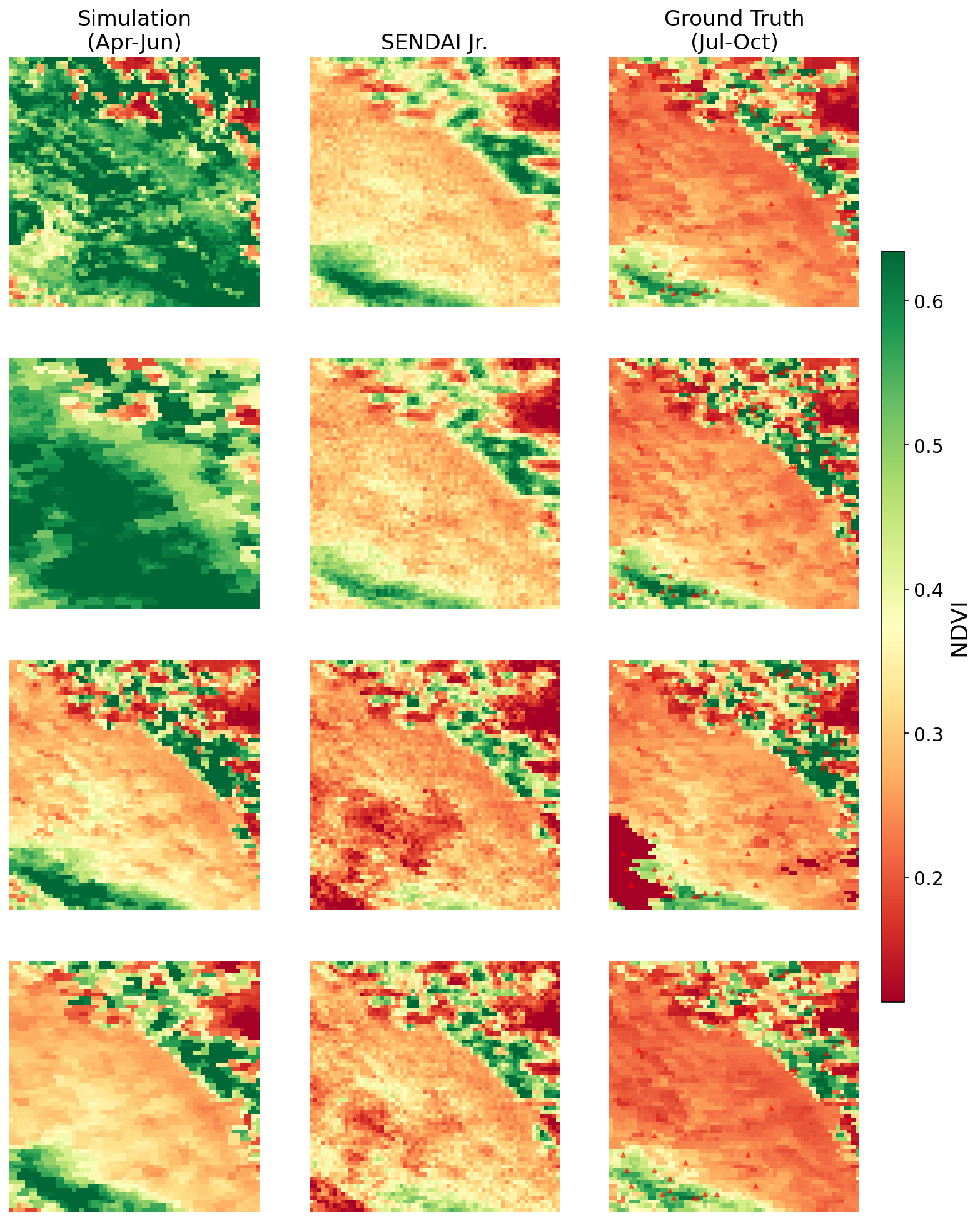}
\caption{SENDAI Jr. reconstruction for Central Valley, California. Each row represents an equally-spaced temporal frame within the ground truth period (July--October). Columns show simulation-period reference (April--June), SENDAI Jr. reconstruction, and ground truth. Red markers indicate sensor locations. The SSIM of 0.5747 reflects the preservation of field-scale spatial structure.}
\label{fig:dashred_cv_app}
\end{figure}

Figure~\ref{fig:dashred_cv_app} presents qualitative reconstruction results across four equally-spaced temporal frames within the ground truth observation period. The reconstructed fields preserve the essential spatial heterogeneity of the ground truth, including field boundaries and vegetation gradients, despite access to only 64 point measurements per frame. The high SSIM value (0.5747) quantitatively confirms this visual assessment of structural preservation.

\subsection{Corn Belt, Iowa, USA}

Despite the pronounced phenological transition from vegetative growth (April--June) to reproductive and senescence stages (July--October), SENDAI Jr. achieves RMSE of 0.1103 and SSIM of 0.4530 at this site. The SSIM improvement is particularly striking: a 185\% increase over the best baseline (SG+IDW at 0.1588). This dramatic improvement reflects SENDAI Jr.'s ability to maintain spatial coherence across the severe phenological domain shift characteristic of temperate agricultural systems.

\begin{figure}[t]
\centering
\includegraphics[width=\columnwidth]{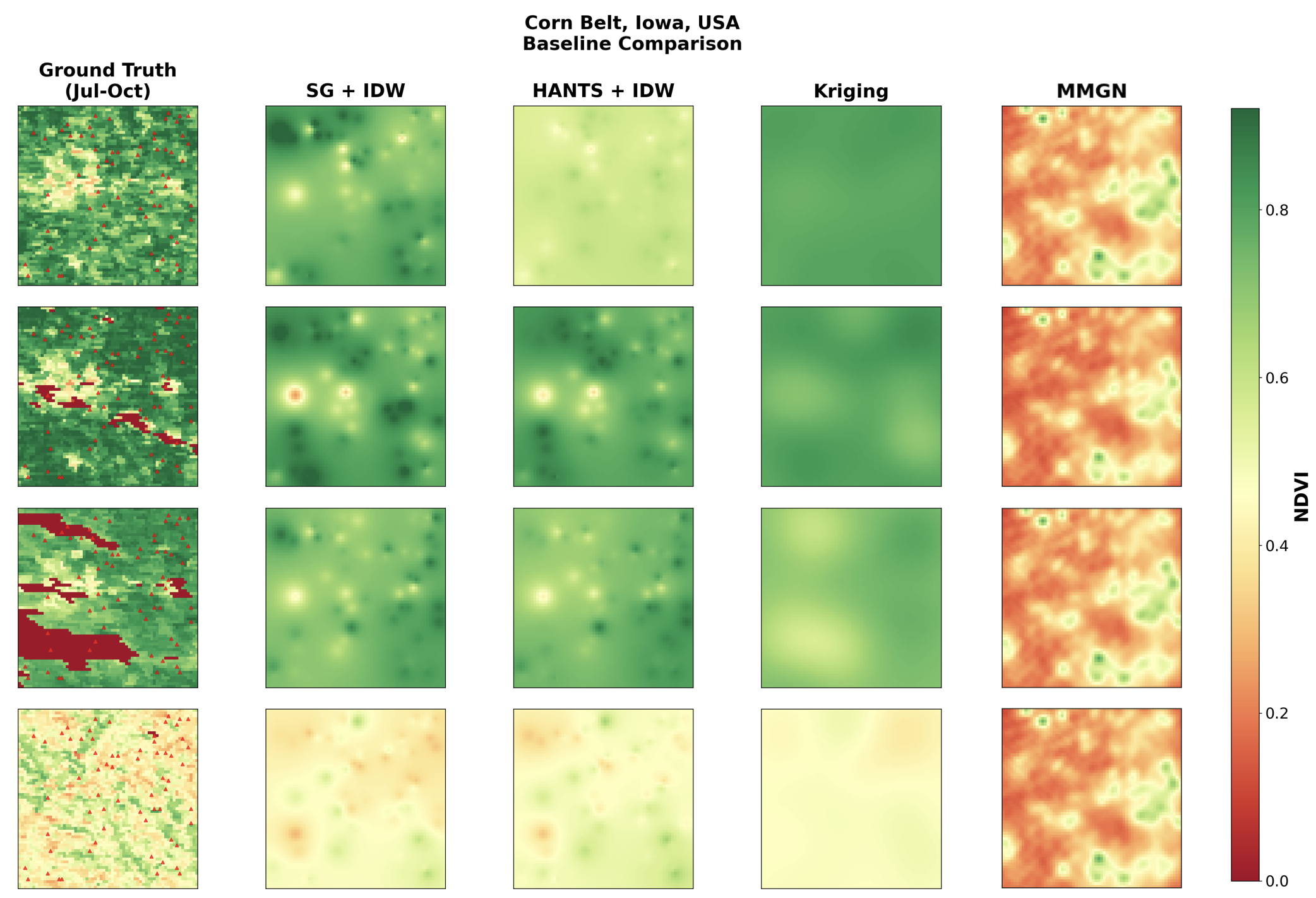}
\caption{Baseline reconstruction comparison for the Iowa Corn Belt. Each row corresponds to an equally-spaced temporal frame during July--October. Despite the relatively homogeneous landscape, baseline methods produce interpolation artifacts that obscure the underlying field structure. Kriging achieves the best RMSE but worst SSIM, demonstrating the importance of structural metrics.}
\label{fig:baseline_iowa_app}
\end{figure}

Figure~\ref{fig:baseline_iowa_app} presents baseline performance on the Iowa site. The rainfed cropland exhibits more homogeneous vegetation patterns than irrigated sites, yet baseline methods still fail to capture the field-scale structure characteristic of corn-soybean rotations. Kriging achieves the lowest SSIM (0.0312), producing nearly featureless smooth fields that completely eliminate the spatial patterns present in the ground truth. This extreme structural degradation occurs despite Kriging achieving the best baseline RMSE (0.1596), highlighting the inadequacy of RMSE as a sole evaluation metric.

\begin{figure}[t]
\centering
\includegraphics[width=\columnwidth]{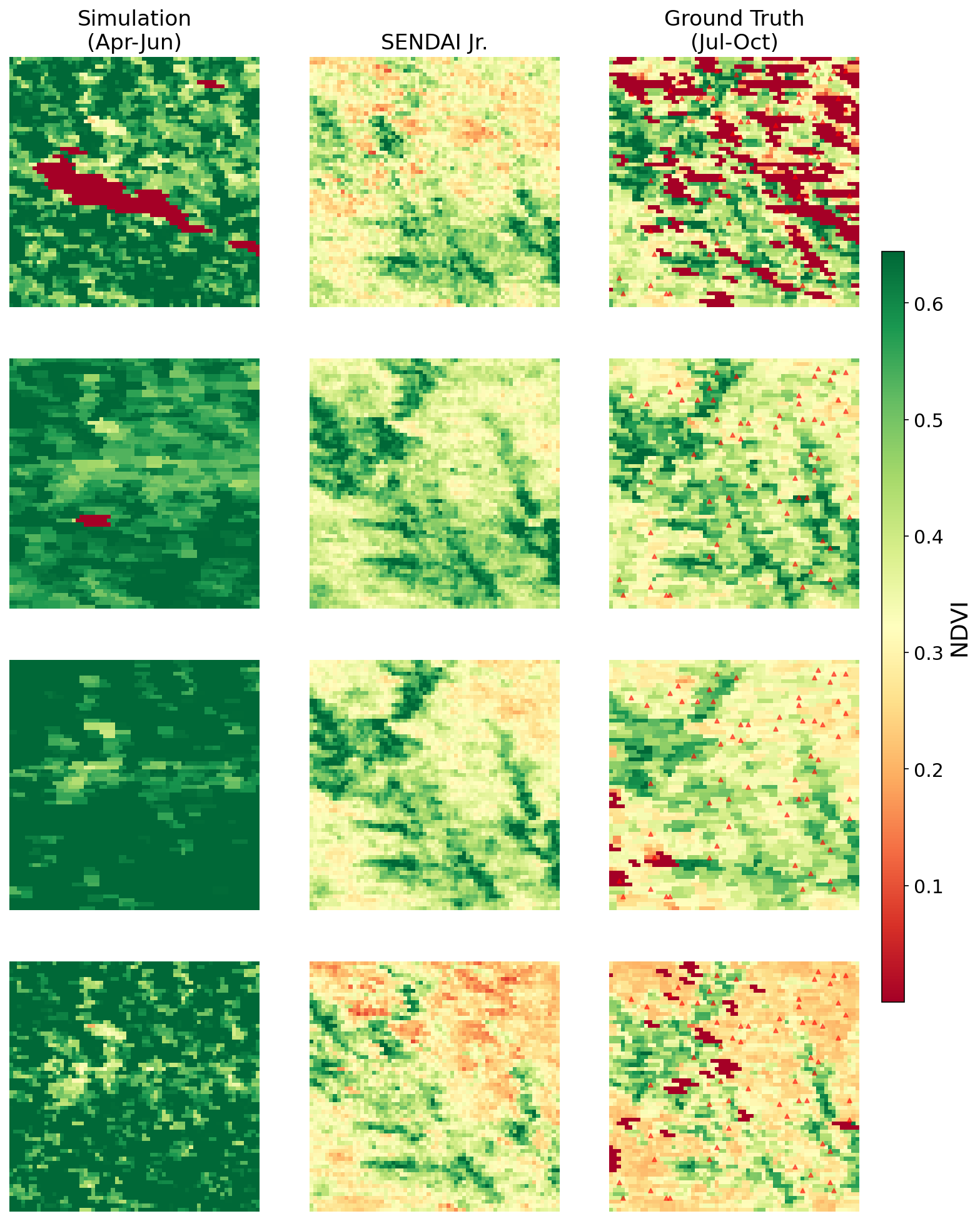}
\caption{SENDAI Jr. reconstruction for the Iowa Corn Belt. Each row corresponds to an equally-spaced temporal frame during July--October. The framework captures the characteristic high-NDVI patterns of mature corn and soybean crops from sparse sensor observations, achieving SSIM of 0.4530.}
\label{fig:dashred_iowa_app}
\end{figure}

Figure~\ref{fig:dashred_iowa_app} illustrates the SENDAI Jr. reconstruction, where the framework successfully adapts from spring emergence conditions to mid-season reproductive stages despite the substantial phenological shift. The reconstruction captures the characteristic high-NDVI patterns of mature corn and soybean crops, preserving the field-scale spatial structure that enables crop type discrimination and yield estimation.

\subsection{Guadalquivir Valley, Spain}

This Mediterranean site, characterized by reversed phenological timing relative to temperate Northern Hemisphere regions, achieves RMSE of 0.1474 and SSIM of 0.3655. The SSIM represents a 98\% improvement over the best baseline (SG+IDW at 0.1849). The framework successfully adapts from winter-spring simulation conditions (February--April) to autumn ground truth observations (September--December), demonstrating robustness to non-standard seasonal calendars.

\begin{figure}[t]
\centering
\includegraphics[width=\columnwidth]{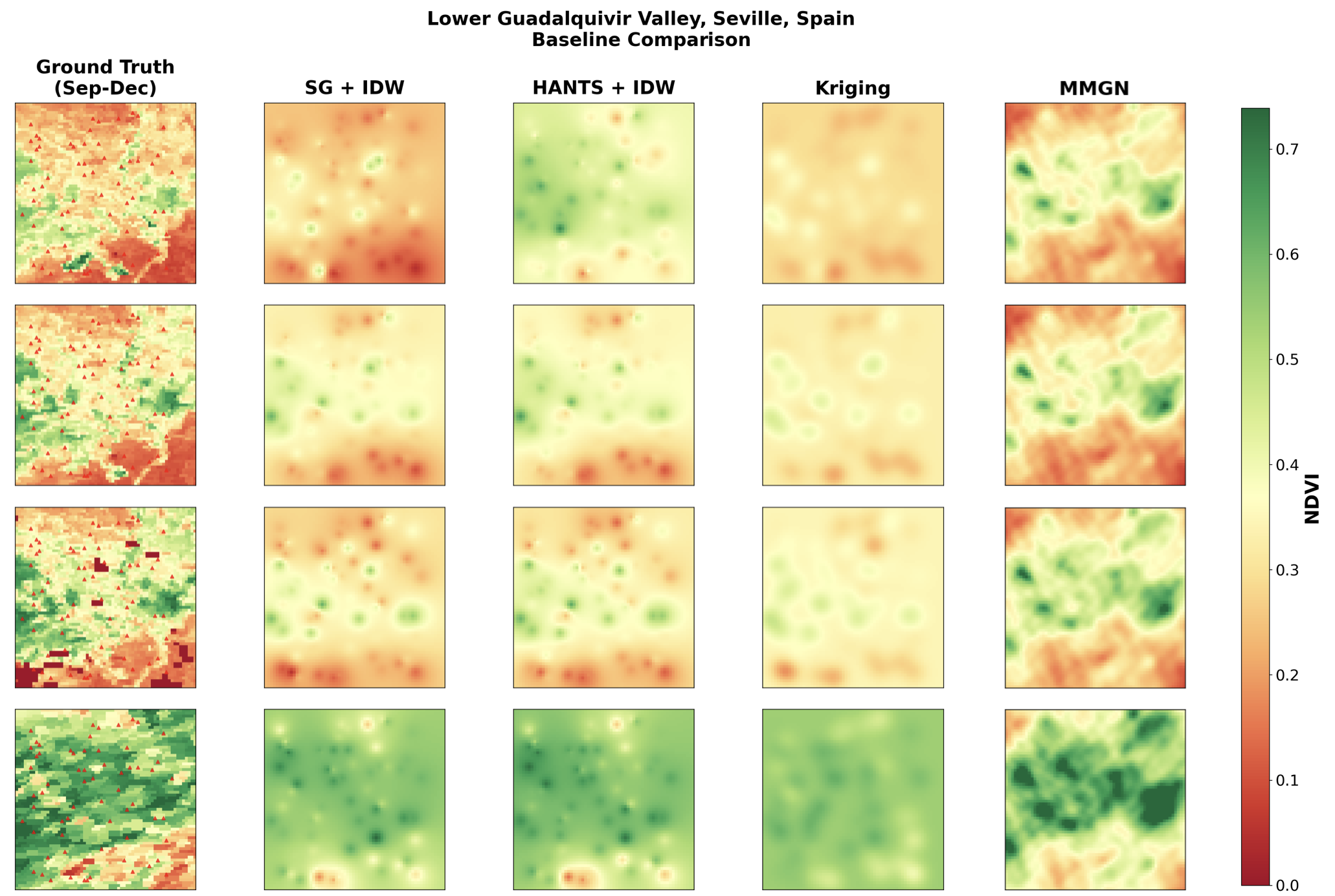}
\caption{Baseline reconstruction comparison for the Guadalquivir Valley, Spain. Each row represents an equally-spaced temporal frame. The mixed agricultural landscape with varying field sizes challenges interpolation-based methods. All baselines achieve SSIM below 0.19, indicating severe structural degradation.}
\label{fig:baseline_spain_app}
\end{figure}

Figure~\ref{fig:baseline_spain_app} shows baseline performance on this site. The mixture of irrigated agriculture and natural vegetation creates heterogeneous patterns that baseline methods fail to faithfully reproduce. All baseline methods achieve SSIM below 0.19, indicating severe structural degradation. Kriging again shows the largest disconnect between RMSE and SSIM, achieving moderate RMSE (0.1481) but poor SSIM (0.0878).

\begin{figure}[t]
\centering
\includegraphics[width=\columnwidth]{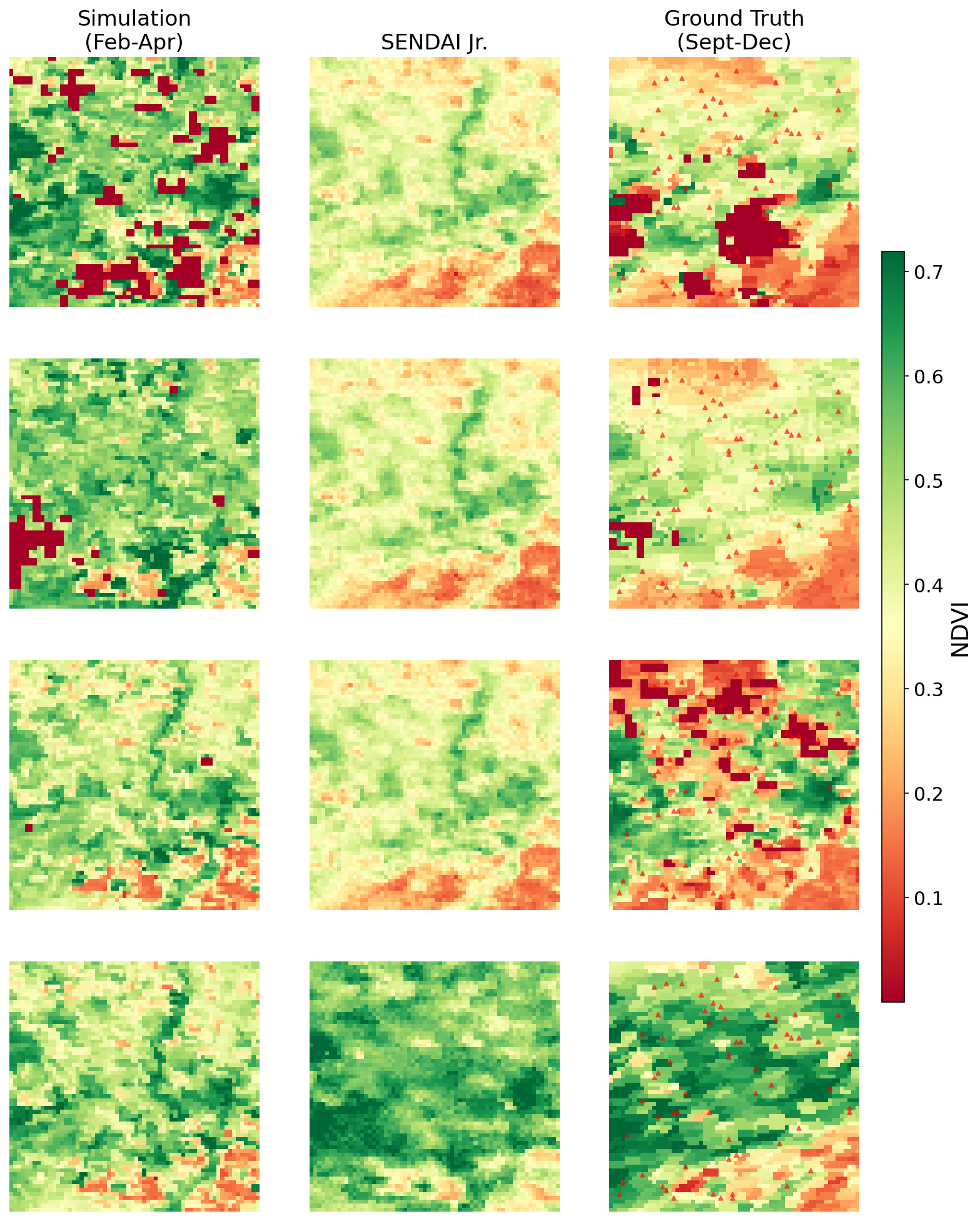}
\caption{SENDAI Jr. reconstruction for the Guadalquivir Valley, Spain. Temporal frames span September--December ground truth observations, reconstructed from February--April simulation training. The model recovers coherent spatial patterns including agricultural field boundaries and regional vegetation gradients, achieving SSIM of 0.3655.}
\label{fig:dashred_spain_app}
\end{figure}

Figure~\ref{fig:dashred_spain_app} demonstrates SENDAI Jr. reconstruction for the Guadalquivir Valley, where the reversed Mediterranean phenological calendar requires adaptation from winter-spring greenness to autumn ground truth conditions. The model recovers coherent spatial patterns including agricultural field boundaries and regional vegetation gradients, as reflected in the substantially improved SSIM.

\subsection{Summary of SENDAI Jr. Performance}

Across all three sites, SENDAI Jr. demonstrates consistent superiority over baseline methods in both RMSE and SSIM metrics, with particularly pronounced advantages in structural preservation:

\begin{itemize}
    \item \textbf{Central Valley}: 120\% SSIM improvement, preserving irrigated field boundaries
    \item \textbf{Iowa Corn Belt}: 185\% SSIM improvement, maintaining field-scale crop patterns
    \item \textbf{Guadalquivir Valley}: 98\% SSIM improvement, capturing Mediterranean agricultural structure
\end{itemize}

The large gap between baseline RMSE and SSIM performance---particularly for Kriging---demonstrates that traditional error metrics inadequately capture reconstruction quality for heterogeneous landscapes. SENDAI Jr.'s success in improving both metrics simultaneously indicates that the framework learns physically meaningful spatial priors rather than simply minimizing point-wise error.

\section{SENDAI Site-Specific Results}
\label{app:multiscale_details}

This appendix provides detailed analyses and qualitative reconstruction results for the three sites evaluated using the full SENDAI hierarchical multiscale DA-SHRED architecture with INR. These sites exhibit complex phenological dynamics, sub-seasonal variability, or pronounced spatial heterogeneity that require high-frequency correction beyond what SENDAI Jr. can provide. SSIM serves as our primary performance indicator, capturing the preservation of sharp boundaries, fine-scale features, and topological structure.

\subsection{Imperial Valley, California, USA}

This dry and hot irrigated agriculture site exhibits consistent improvement through the hierarchical pipeline. The Cheap2Rich achieves SSIM of 0.4041, which increases to 0.4411 with HF peeling (+9.2\%) and further to 0.4668 with the full SENDAI pipeline (+15.5\% total improvement from Cheap2Rich). RMSE improves correspondingly from 0.1708 to 0.1486.

The baseline methods achieve uniformly poor SSIM values: SG+IDW (0.1123), HANTS+IDW (0.1049), and Kriging (0.0916). This represents a 4$\times$ to 5$\times$ improvement in structural similarity for the full SENDAI pipeline over baselines, demonstrating the framework's ability to preserve the rectilinear irrigation infrastructure that defines this landscape.

\begin{figure}[t]
\centering
\includegraphics[width=\columnwidth]{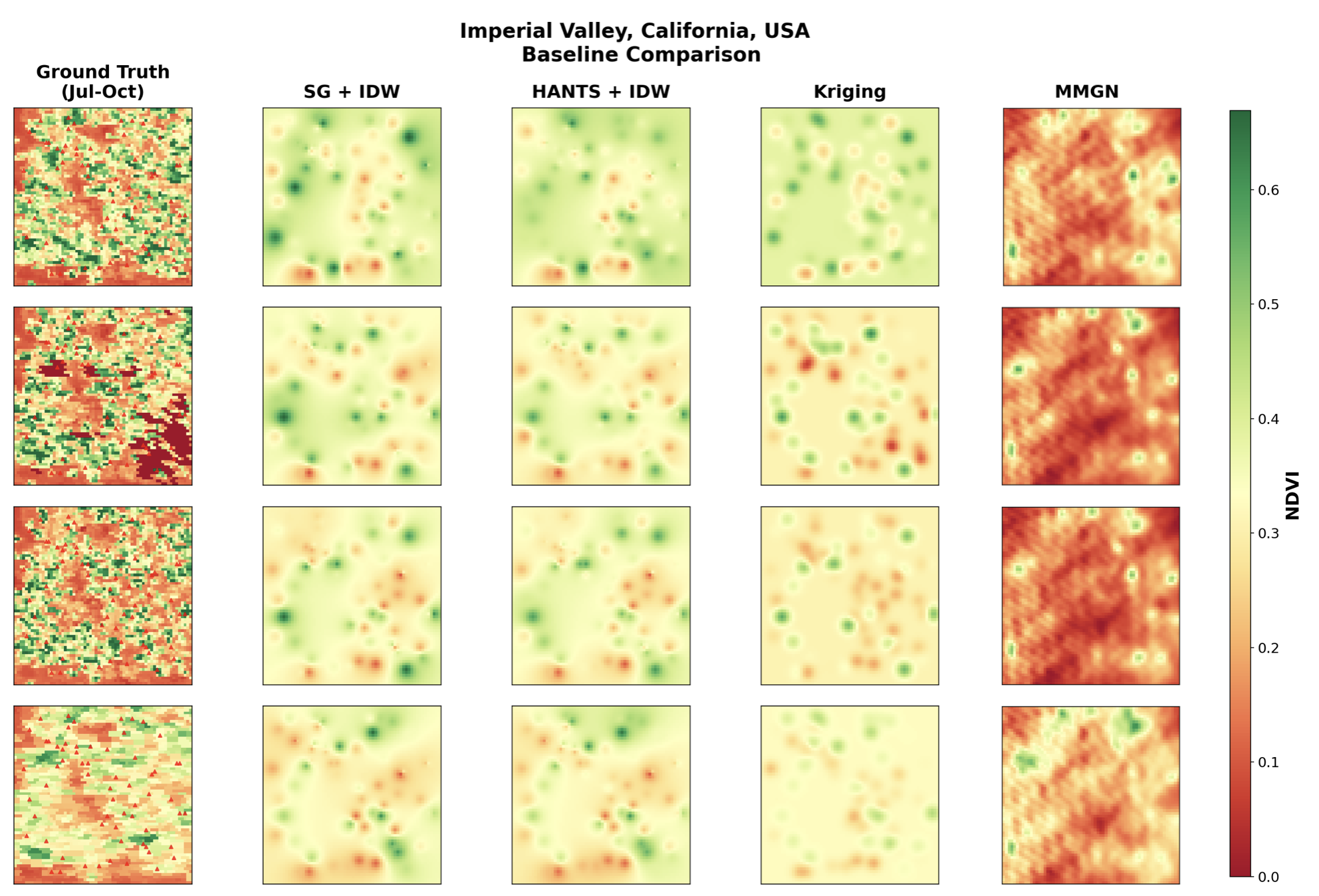}
\caption{Baseline reconstruction comparison for Imperial Valley, California. Each row corresponds to an equally-spaced temporal frame. The rectilinear irrigation infrastructure creates sharp NDVI boundaries that baseline methods fail to preserve, producing either circular artifacts or smooth gradients. Baseline SSIM values range from 0.0916 to 0.1123.}
\label{fig:baseline_iv_app}
\end{figure}

Figure~\ref{fig:baseline_iv_app} presents baseline reconstruction results for Imperial Valley. The contrast between irrigated fields and surrounding desert creates sharp NDVI boundaries that are lost in baseline reconstructions, manifesting as circular bullseye artifacts (IDW methods) or smooth gradients (Kriging). The low SSIM values quantify this structural failure.

\begin{figure}[t]
\centering
\includegraphics[width=\columnwidth]{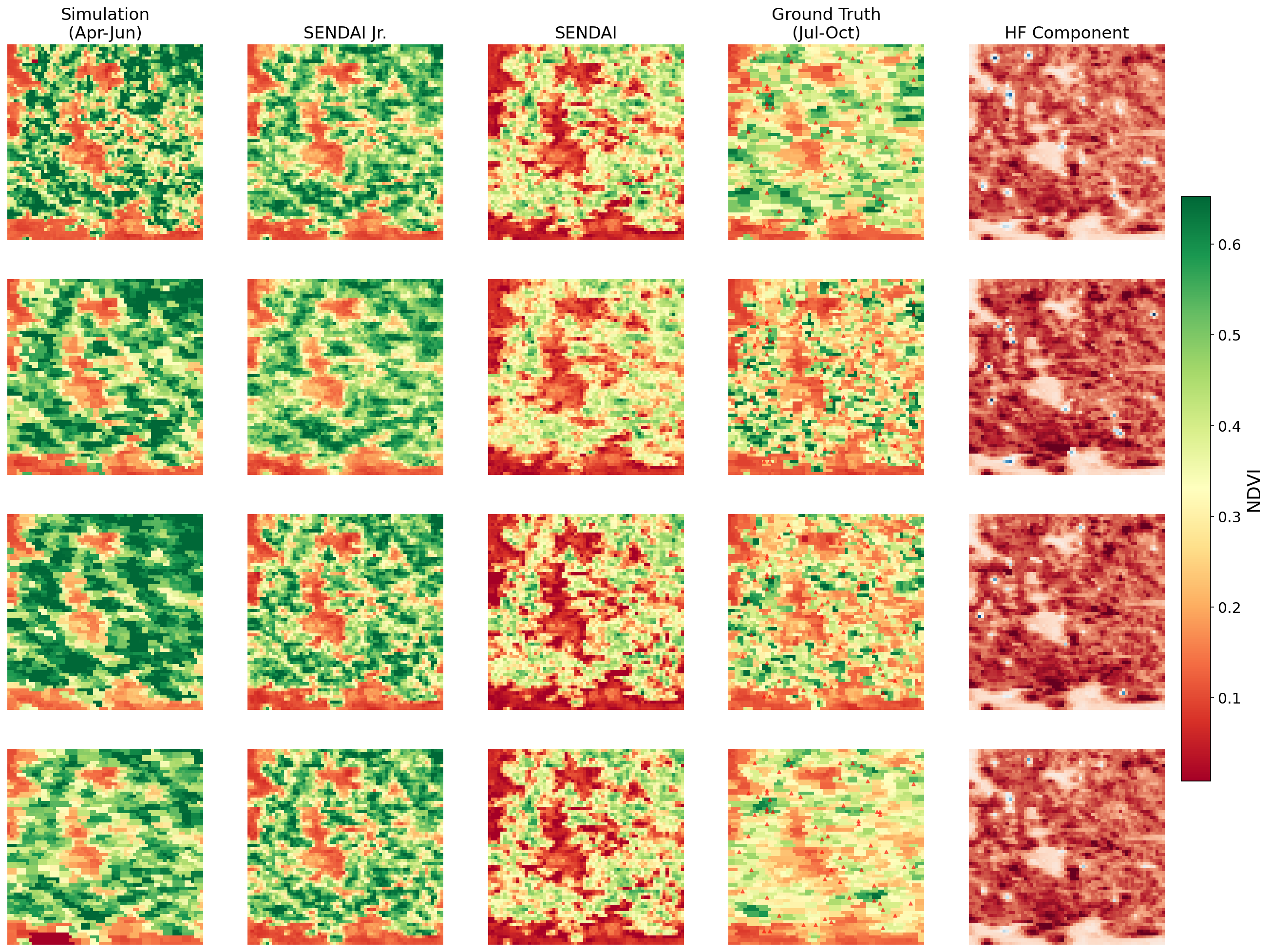}
\caption{Full Sendai hierarchical multi-scale DA-SHRED reconstruction for Imperial Valley, California. Each row corresponds to an equally-spaced temporal frame during July--October. Columns show: simulation reference (April--June), SENDAI Jr. stage, full SENDAI hierarchical reconstruction, ground truth, and the learned HF correction component. SENDAI captures fine-scale field boundaries, achieving SSIM of 0.4668 in the full pipeline.}
\label{fig:multiscale_iv_app}
\end{figure}

Figure~\ref{fig:multiscale_iv_app} shows the full hierarchical reconstruction pipeline. The HF pathway captures fine-scale field boundaries characteristic of the rectilinear irrigation infrastructure, with the learned HF component exhibiting coherent spatial structure aligned with field edges rather than random noise.

\subsection{Tarim Basin, China}

This continental site presents the most pronounced contrast between Cheap2Rich and Sendai performance. The Cheap2Rich achieves SSIM of 0.3505, which increases to 0.4257 with HF peeling (+21.5\%) and to 0.4777 with the full pipeline (+36.3\% total improvement). This is the largest SSIM improvement across all six sites, reflecting the critical importance of hierarchical high-frequency correction for landscapes with sharp boundaries.

The hierarchical pipeline also achieves substantial RMSE improvement: from 0.1827 (Cheap2Rich) to 0.1208 (SENDAI), a 33.9\% reduction. This demonstrates that the high-frequency corrections are not merely cosmetic but represent genuine improvements in reconstruction accuracy.

Baseline methods fail catastrophically on this site in terms of structural preservation: Kriging achieves the worst SSIM (0.0449), producing nearly featureless smooth fields despite moderate RMSE. The sharp mountain-basin boundaries that define this landscape are completely unresolved by all baseline approaches.

\begin{figure}[t]
\centering
\includegraphics[width=\columnwidth]{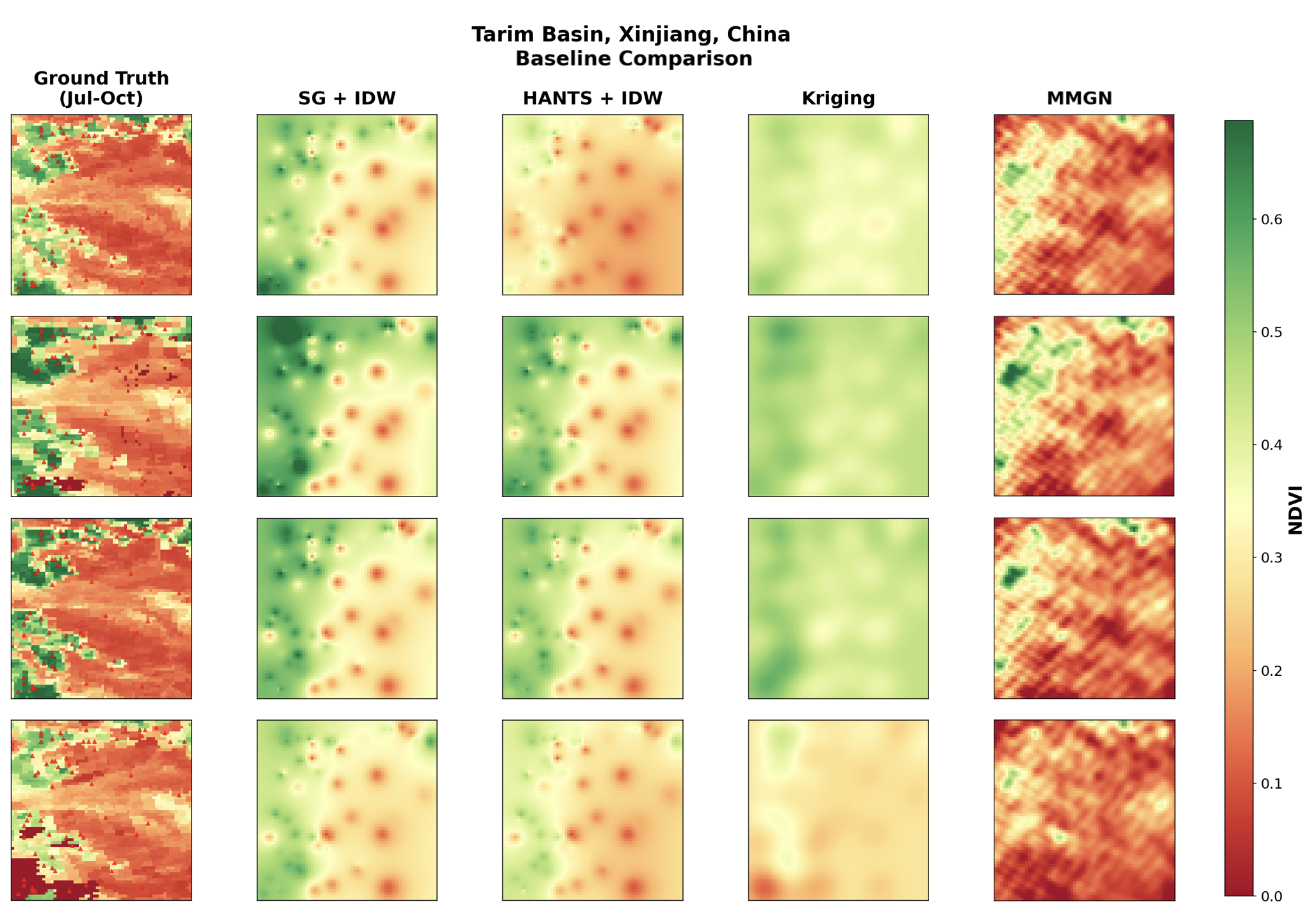}
\caption{Baseline reconstruction comparison for the Tarim Basin, Xinjiang, China. Each row represents an equally-spaced temporal frame. The sharp mountain-basin boundaries that define this landscape are completely lost in all baseline reconstructions, demonstrating the fundamental limitations of interpolation-based approaches for heterogeneous terrain. Kriging achieves SSIM of only 0.0449.}
\label{fig:baseline_xj_app}
\end{figure}

Figure~\ref{fig:baseline_xj_app} illustrates the failure of baseline methods on this challenging site. The oasis-desert transition creates extreme spatial gradients that interpolation-based methods cannot capture. IDW methods produce pronounced bullseye artifacts, while Kriging eliminates all boundary information.

\begin{figure}[t]
\centering
\includegraphics[width=\columnwidth]{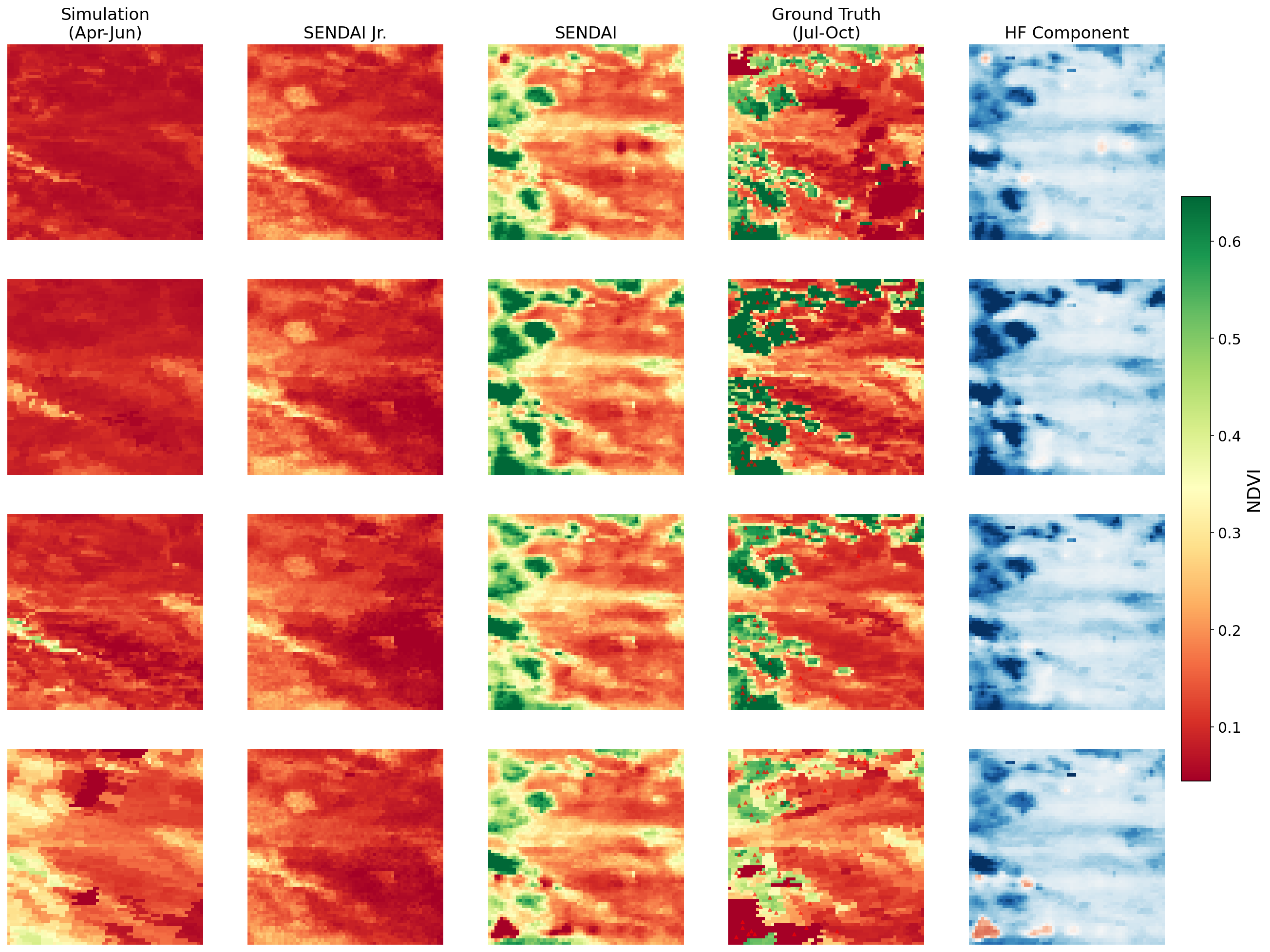}
\caption{Full Sendai hierarchical multi-scale DA-SHRED reconstruction for the Tarim Basin, Xinjiang, China. Temporal frames span July--October ground truth observations. The HF component captures sharp oasis-desert boundaries that the smooth LF decoder fails to resolve, with the full pipeline achieving SSIM of 0.4777 and 33.9\% RMSE improvement over Cheap2Rich.}
\label{fig:multiscale_xj_app}
\end{figure}

Figure~\ref{fig:multiscale_xj_app} presents the full hierarchical reconstruction pipeline. The HF component captures the sharp oasis-desert boundaries that the smooth LF decoder fails to resolve. The learned corrections exhibit spatially coherent structure aligned with the landscape's topological features, confirming that the peeling layers discover physically meaningful high-frequency content rather than noise.

\subsection{Riverina, Australia}

This mixed cropping region exhibits SSIM improvement from 0.2761 (Cheap2Rich) to 0.3158 (Cheap2Rich+HFP, +14.4\%) and 0.3354 (SENDAI, +21.5\%). While the absolute SSIM values are lower than the other sites, this reflects the inherently more diffuse spatial structure of this landscape rather than reconstruction failure.

Baseline methods again show poor structural preservation: Kriging achieves SSIM of only 0.0272, the lowest across all six sites. Even SG+IDW and HANTS+IDW achieve only 0.1359 and 0.1245 respectively. The full SENDAI pipeline achieves 2.5$\times$ higher SSIM than the best baseline.

\begin{figure}[t]
\centering
\includegraphics[width=\columnwidth]{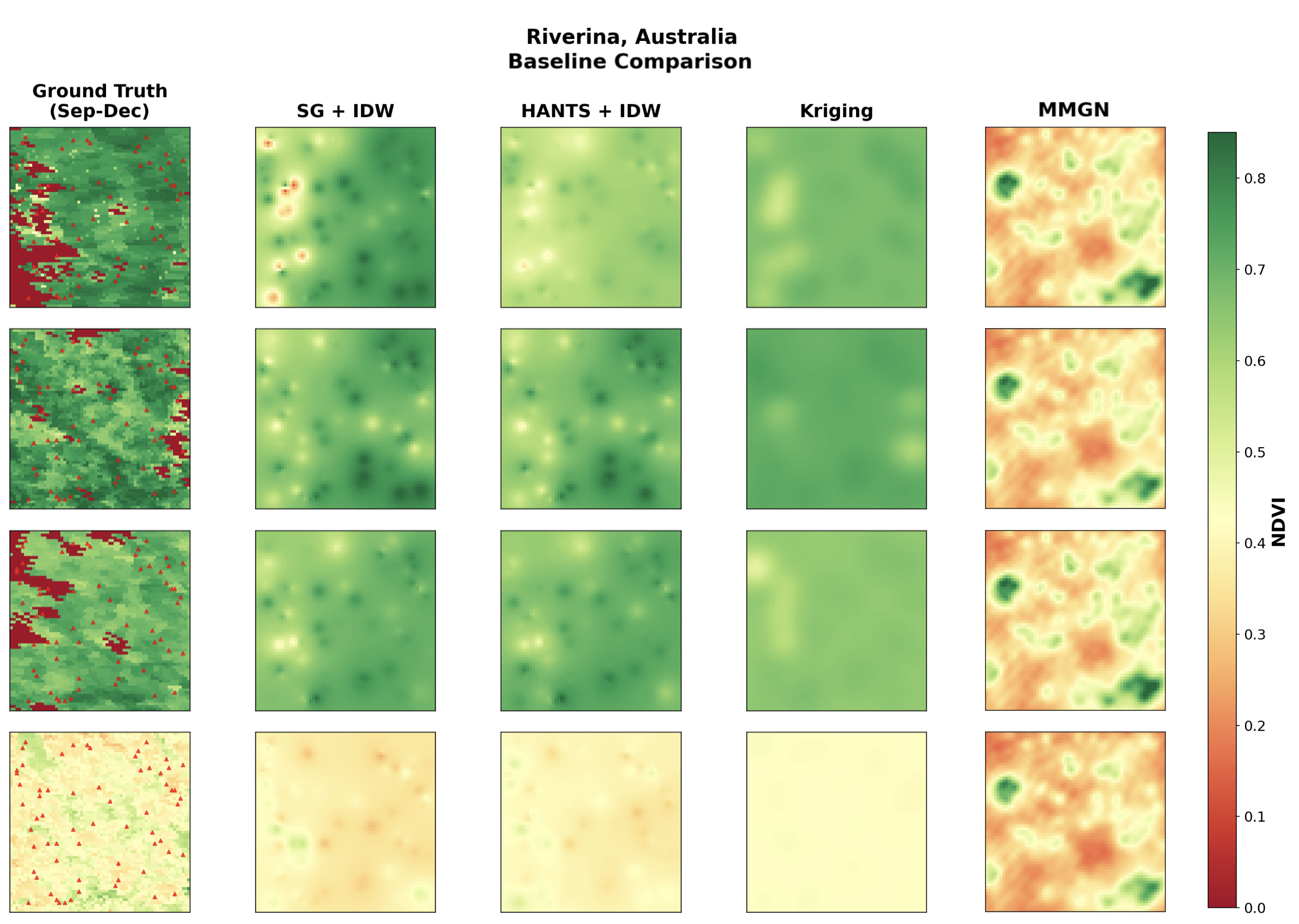}
\caption{Baseline reconstruction comparison for Riverina, Australia. Each row represents an equally-spaced temporal frame. The mixed cropping landscape exhibits more gradual spatial transitions than other sites, but baseline methods still produce artifacts and over-smoothing. Baseline SSIM values range from 0.0272 to 0.1359.}
\label{fig:baseline_aus_app}
\end{figure}

Figure~\ref{fig:baseline_aus_app} shows baseline performance on the Riverina site. While the spatial heterogeneity is less extreme than at other sites, baseline methods still fail to capture the field-scale structure of this mixed cropping landscape.

\begin{figure}[t]
\centering
\includegraphics[width=\columnwidth]{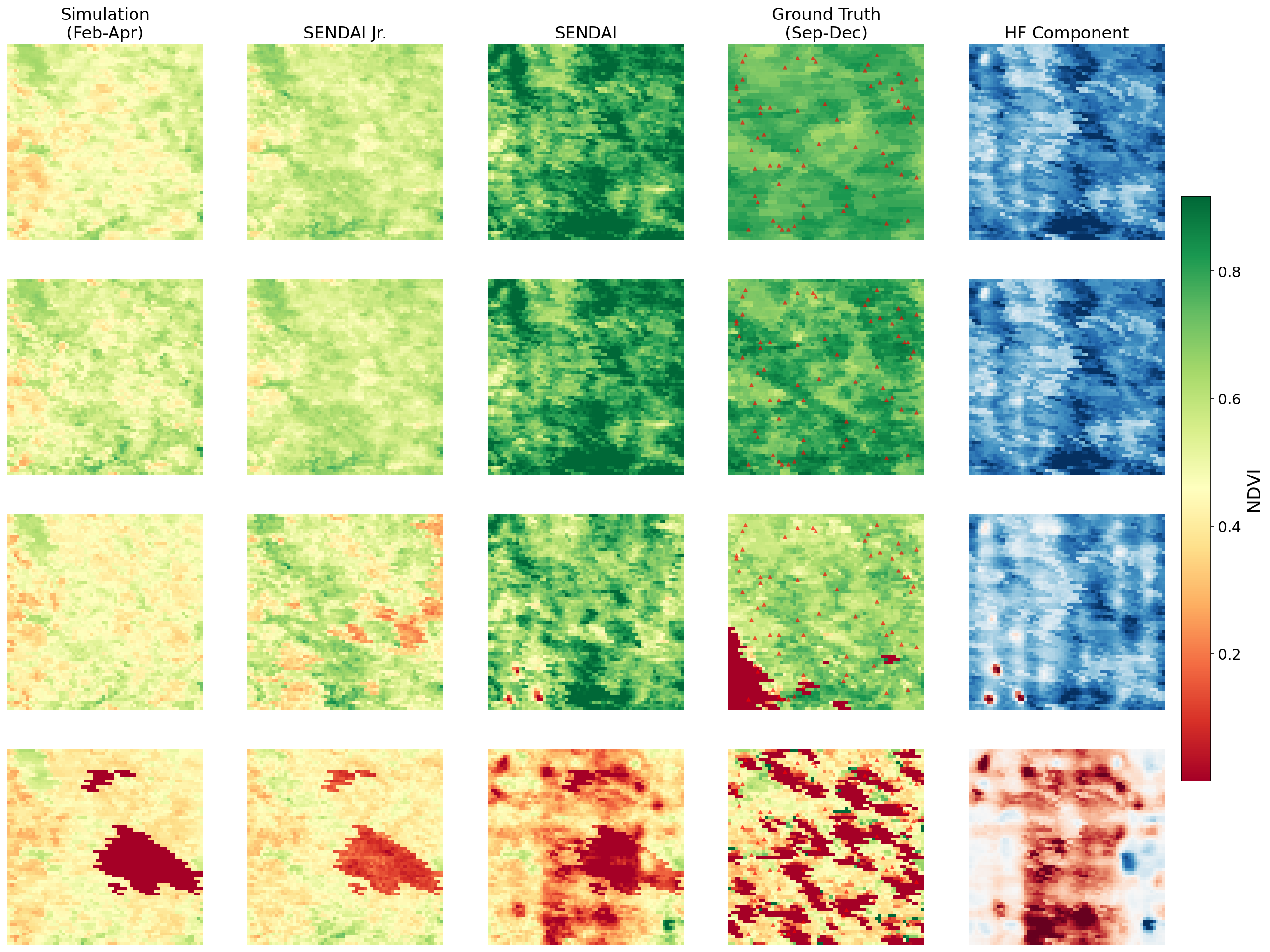}
\caption{Full SENDAI hierarchical multi-scale DA-SHRED reconstruction for Riverina, Australia. Each row represents an equally-spaced temporal frame during the September--December ground truth period. The Southern Hemisphere site tests generalization to reversed seasonal timing, with the HF component showing more diffuse structure consistent with the gradual spatial transitions in this mixed cropping region. Full pipeline achieves SSIM of 0.3354.}
\label{fig:multiscale_aus_app}
\end{figure}

Figure~\ref{fig:multiscale_aus_app} presents the hierarchical reconstruction. The HF component shows more diffuse structure consistent with the gradual spatial transitions in this mixed cropping region, reflecting the site's inherent landscape characteristics. The Southern Hemisphere location provides a test of generalization to reversed seasonal timing.

\subsection{Summary of SENDAI Performance}

The hierarchical SENDAI framework demonstrates consistent improvements over both baselines and Cheap2Rich across all three challenging sites:

\begin{itemize}
    \item \textbf{Imperial Valley}: 15.5\% SSIM improvement from Cheap2Rich, 4.2$\times$ improvement over best baseline
    \item \textbf{Tarim Basin}: 36.3\% SSIM improvement from Cheap2Rich, 3.7$\times$ improvement over best baseline  
    \item \textbf{Riverina}: 21.5\% SSIM improvement from Cheap2Rich, 2.5$\times$ improvement over best baseline
\end{itemize}

\subsection{Sensor Number Sensitivity Analysis}
\label{appendix:sensor_sensitivity}

We conducted a sensitivity analysis to evaluate the effect of sensor density on reconstruction quality. Experiments were performed with sensor counts ranging from 8 to 256, with five independent trials per configuration.

Figure~\ref{fig:dashred_cv_main}a shows SSIM performance as a function of sensor count. Both the full dataset and validation set exhibit a clear positive trend: reconstruction quality improves with increasing sensor density. For our main experiments, we selected 64 sensors as a conservative configuration. At this density, the model achieves mean SSIM values of approximately 0.53 (full) and 0.49 (validation), representing a reasonable trade-off between reconstruction accuracy and practical deployment constraints. While higher sensor counts yield improved performance, the marginal gains must be weighed against increased instrumentation costs and maintenance requirements in field applications.


\section{Hardware Efficiency: Extended Discussion}
\label{app:hardware_efficiency_details}

This appendix provides detailed operational scenarios for the hardware efficiency paradigm introduced in Section~\ref{sec:results_efficiency}.

The demonstrated capacity to reconstruct full fields from 64 sensors (1.56\% of pixels) suggests alternative paradigms for satellite data systems. Rather than transmitting complete imagery, systems could transmit sparse measurements alongside periodically updated model weights, achieving substantial data reduction while preserving spatial structure essential for downstream analysis. This paradigm addresses bandwidth constraints in resource-limited regions~\citep{de2011reliability}, onboard storage limitations, and low-latency decision support requirements where provisional reconstructions from partial observations enable time-critical responses~\citep{denby2020orbital, giuffrida2020cloudscout}.

\subsection{Bandwidth Reduction Scenario}

Consider the operational scenario: a satellite acquires a $64 \times 64$ pixel scene (4,096 values). Under our sparse sensing protocol, only 64 sensor measurements need be transmitted---a 64$\times$ reduction before any conventional compression. If model weights are updated infrequently (e.g., seasonally), the marginal transmission cost per scene becomes negligible. While reconstruction fidelity will never match lossless transmission of complete imagery, for applications where the reconstructed fields serve as inputs to downstream models (crop monitoring, anomaly detection, change analysis), the accuracy-bandwidth tradeoff may prove favorable.

\subsection{Deep-Space Exploration Applications}

This paradigm holds particular significance for deep-space exploration missions, where communication bandwidth constrains scientific return~\citep{de2011reliability, xie2021compression}. Planetary surface monitoring---whether for Mars rover operations, lunar resource mapping, or asteroid characterization---faces extreme bandwidth limitations. A SENDAI-like architecture, pretrained on simulation data or initial mission observations, could enable substantial compression of subsequent observations while preserving the spatial structure essential for scientific interpretation.

\subsection{Low-Latency Decision Support Scenario}

Consider a scenario where only a subset of spectral bands has completed atmospheric correction, or where preliminary telemetry provides sensor-location values before full-scene processing completes. The framework can generate an immediate reconstruction that, while provisional, may suffice for time-critical decisions. Full-fidelity imagery, when available, supersedes the provisional estimate.

This capability aligns with emerging architectures for on-board satellite processing~\citep{denby2020orbital, giuffrida2020cloudscout}, where computational constraints preclude comprehensive analysis but rapid triage decisions (e.g., assessing scenes to be prioritized for downlink) could leverage lightweight reconstruction from sparse measurements.

\section{Domain Adaptation: Extended Discussion}
\label{app:domain_adaptation_details}

This appendix provides extended discussion of domain adaptation applications introduced in Section~\ref{sec:conclusions}.

\subsection{Application to Soil Moisture Retrieval}

Consider soil moisture retrieval, where physics-based radiative transfer models provide simulation data but real-world observations exhibit systematic biases from surface roughness, vegetation attenuation, and instrument calibration~\citep{mohanty2017soil}. The SENDAI framework provides a principled approach to bridge such gaps: pretrain on simulation or reference-period data to establish spatial priors, then align to target conditions using sparse real observations from in-situ networks.

\subsection{Application to Land Surface Temperature}

Similarly, land surface temperature products exhibit seasonal and diurnal biases that complicate multi-temporal analysis~\citep{li2013satellite}. Physics-based energy balance models can provide simulation priors, while sparse thermal observations from clear-sky periods serve as target-domain data for latent-space alignment.

\subsection{Implications for Operational Monitoring}

The implications for operational monitoring systems are substantial. Near-real-time agricultural monitoring---critical for yield forecasting, drought assessment, and food security early warning~\citep{gao2021mapping, weiss2020remote}---requires rapid processing of incoming satellite data streams. Current systems face a fundamental tension: sophisticated reconstruction methods improve data quality but introduce latency; rapid processing preserves timeliness but sacrifices accuracy. SENDAI's short inference times per scene suggest a viable path toward high-quality reconstruction within operational latency constraints.

Furthermore, the framework's ability to reconstruct from sparse observations has implications for monitoring in resource-constrained settings. Regions with limited ground-station coverage, intermittent connectivity, or bandwidth constraints---common in developing nations and remote areas---could potentially achieve comparable monitoring fidelity by transmitting only sparse sensor measurements alongside periodically updated model parameters, rather than full imagery.

\section{Implicit Neural Representations for Geospatial Data}
\label{app:inr}

The coordinate-based implicit neural representation (INR) employed in the high-frequency pathway represents an emerging paradigm for geospatial data with substantial unexplored potential~\citep{sitzmann2020implicit, tancik2020fourier}. Unlike discrete gridded representations, INRs parameterize spatial fields as continuous functions, enabling queries at arbitrary coordinates and natural handling of irregular geometries.

Recent work has demonstrated INR applicability across Earth science domains: potential field geophysics~\citep{smith2025implicit}, species distribution modeling~\citep{cole2023spatial}, and climate data compression~\citep{mostajeran2025context}. Our application to high-frequency correction in satellite imagery suggests additional utility: INRs can learn spatially coherent patterns from sparse sensor residuals, producing smooth interpolation without the localized artifacts characteristic of direct MLP regression.

The combination of INR decoders with recurrent encoders of sensor time-histories represents a hybrid architecture with broader applicability. The recurrent encoder captures temporal dynamics (phenological trends, seasonal patterns); the INR decoder produces spatially coherent instantaneous fields. This separation of temporal and spatial processing may prove advantageous across spatiotemporal reconstruction problems where temporal and spatial structure exhibit distinct characteristics.

\end{document}